\documentclass[12pt]{article}

\usepackage{fullpage}

\usepackage{placeins}

\usepackage{cite}
\usepackage{graphicx}

\usepackage{amsmath, amssymb, amsthm}
\usepackage{mathtools}

\usepackage[ruled,linesnumbered,noend]{algorithm2e}
\usepackage{algorithmic}

\usepackage[printonlyused]{acronym}

\usepackage{tabularx}
\usepackage{multirow}
\usepackage{hhline}
\usepackage{makecell}

\usepackage{pbox}

\usepackage{array}
\renewcommand{\arraystretch}{1.2}
\newcolumntype{C}[1]{>{\centering\arraybackslash}m{#1}}
\usepackage{arydshln} 

\usepackage[font=scriptsize]{subcaption}

\usepackage{stfloats} 

\usepackage{url}

\usepackage{tikz}
\usepackage{pgf}
\usepackage{pgfplots}
\usepackage{color}
\usepackage{fp}
\usetikzlibrary{arrows,automata,patterns,backgrounds,shapes} 
\usepgfplotslibrary{external} 
\pgfplotsset{compat=1.14}

\usepackage{bm}
\SetSymbolFont{largesymbols}{bold}{OMX}{txex}{b}{n}

\usepackage{cancel}
\usepackage[normalem]{ulem}

\usepackage{pifont}

\usepackage{xspace}
\newcommand{\ie}{\mbox{i.\,e.,}\xspace}

\DeclareMathAlphabet{\mathpzc}{OT1}{pzc}{m}{it}

\usepackage{xifthen}

\newcommand{\R}{ {\mathbb{R}} }

\newcommand{\E}[1]{E\left[{#1}\right]}
\newcommand{\Var}[1]{{\it Var}\left[{#1}\right]}
\newcommand{\EEst}[1]{\hat{E}\left[{#1}\right]}
\newcommand{\VarEst}[1]{\hat{\it Var}\left[{#1}\right]}
\newcommand{\sigmaEst}[1]{\hat{\sigma}\left[{#1}\right]}
\newcommand{\q}[2]{q_{#1}^{#2}}
\newcommand{\CDF}[2]{F_{#2}\left({#1}\right)}
\newcommand{\UniformNoPar}{\mathcal{U}}
\newcommand{\Uniform}[2]{ {\UniformNoPar}{\left({#1},{#2}\right)}}
\newcommand{\GaussNoPar}{\mathcal{N}}
\newcommand{\Gauss}[2]{ {\GaussNoPar}{\left({#1},{#2}\right)}}
\newcommand{\BetaNoPar}{\mathcal{B}}
\newcommand{\BetaDistribution}[2]{ {\BetaNoPar}{\left({#1},{#2}\right)}}
\newcommand{\hi}[1]{%
  \ifthenelse{\isempty{#1}}{ {\it h} }{ {\it h}_{#1}}%
}
\newcommand{\lo}[1]{%
  \ifthenelse{\isempty{#1}}{ {\it l} }{ {\it l}_{#1}}%
}
\newcommand{\hiEst}[1]{%
  \ifthenelse{\isempty{#1}}{ \overline{\it h} }{ \overline{\it h}_{#1}}%
}
\newcommand{\loEst}[1]{%
  \ifthenelse{\isempty{#1}}{ \overline{\it l} }{ \overline{\it l}_{#1}}%
}
\newcommand{\APW}{C_{\it pw}}
\newcommand{\AUniform}{C_{\it uni}}
\newcommand{\AGauss}{C_{\it{gauss}}}
\newcommand{\AEmpirical}{C_{\it emp}}
\newcommand{\AHistogram}[1]{C_{\it hist}^{#1}}
\newcommand{\AReduced}{C_{\it reduce}}
\newcommand{\AThreeStage}{C_{\it ThreeStage}}
\newcommand{\AMeanBased}{C_{\it Mean}}

\newcommand{\ePW}{e_{\it pw}}

\DeclareMathOperator\erf{erf}

\definecolor{lightgreen}{rgb}{0.8,1.0,0.8}
\definecolor{darkgreen}{rgb}{0.0,0.5,0.125}

\newcommand{\cmark}{\ding{51}}%
\newcommand{\xmark}{\ding{55}}%

\newcommand{\obj}{f}

\definecolor{azure}{rgb}{.0,0.45,0.97}
\definecolor{lightblue}{rgb}{.8,.8,1.0}
\definecolor{midblue}{rgb}{.5,.5,1.0}
\definecolor{lightgreen}{rgb}{0.8,1.0,0.8}
\definecolor{darkgreen}{rgb}{0.0,0.5,0.125}
\definecolor{midgreen}{rgb}{0.0,0.75,0.125}
\definecolor{darkred}{rgb}{0.8,0.2,0.0}
\definecolor{midred}{rgb}{1,0.5,0.5}
\definecolor{lightred}{rgb}{1,0.8,0.8}
\definecolor{darkorange}{rgb}{1,0.35,0.0}
\definecolor{lightorange}{rgb}{1,0.7,0.4}
\definecolor{darkbrown}{rgb}{0.43,0.3,0.16}
\definecolor{lightbrown}{rgb}{0.93,0.86,0.78}

\newcommand*\GnuplotDefs{
    set samples 271;
    Binv(p,q)=exp(lgamma(p+q)-lgamma(p)-lgamma(q));
    beta(x,p,q)=p<=0||q<=0?1/0:x<0||x>1?0.0:Binv(p,q)*x**(p-1.0)*(1.0-x)**(q-1.0);
    normaldistribution(x,m,v)=1/sqrt(2*pi*v)*exp( - ( (x-m)**2)/ (2*v) );
}
 
\newcommand{\RANDVARONE}{X_1}
\newcommand{\RANDVARTWO}{X_2}
\tikzset{%
  SCOPEPROPERTIESDISTRIBUTION/.style={%
    scale=.5, font=\large, xshift=0cm, yshift=0cm
  }%
}

\newcommand{\AXISOPTIONHEIGHT}{7.2cm}

\newcommand{\AXISOPTIONWIDTH}{9.0cm}
\newcommand{\distI}{
\begin{tikzpicture}

	\begin{scope}[SCOPEPROPERTIESDISTRIBUTION]
	    \begin{axis}[height=\AXISOPTIONHEIGHT, width=\AXISOPTIONWIDTH, grid=major, grid style={line width=.1pt, draw=gray!20}, xticklabel style={yshift=-1pt}, xlabel={random variable}, ylabel={probability density}, ylabel style={yshift=2pt}, xlabel style={yshift=-0.5pt}, xmin= 0.0, ymin=0.0, xmax = 1.0, ymax = 7.5, domain=0:1, no markers]
          \addplot gnuplot[raw gnuplot, color=darkgreen, line width=1pt] {
            \GnuplotDefs
            plot [x=0:1] beta(x,4,2) with lines;
          };
          \addplot gnuplot[raw gnuplot, color=darkred, line width=1pt, dashed] {
            \GnuplotDefs
            plot [x=0:1] beta(x,20,20) with lines;
          };
		\legend{{$\RANDVARONE \sim \mathcal{B}(4,2)$},{$\RANDVARTWO \sim \mathcal{B}(20,20)$}}; 
	    \end{axis}     
	\end{scope}

\end{tikzpicture}  
}
\newcommand{\distII}{
\begin{tikzpicture}

	\begin{scope}[SCOPEPROPERTIESDISTRIBUTION] 
	    \begin{axis}[height=\AXISOPTIONHEIGHT, width=\AXISOPTIONWIDTH, grid=major, grid style={line width=.1pt, draw=gray!20}, xticklabel style={yshift=-1pt}, xlabel={random variable}, ylabel={probability density}, ylabel style={yshift=2pt}, xlabel style={yshift=-0.5pt}, xmin= 0.0, ymin=0.0, xmax = 1.0, ymax = 7.5]

		\addplot[mark=none, color=darkgreen, line width=1pt] coordinates { 
			(0.0,0.02)(0.3,0.02)(0.3,2.0)(0.8,2.0)(0.8,0.02)(1.0,0.02)
		}; 
		
		\addplot[mark=none, color=darkred, line width=1pt, dashed] coordinates { 
			(0.0,0.0)(0.4,0.0)(0.4,5.0)(0.6,5.0)(0.6,0.0)(1.0,0.0)
		};
		
		\legend{{$\RANDVARONE \sim \mathcal{U}(0.3,0.8)$},{$\RANDVARTWO \sim \mathcal{U}(0.4,0.6)$}}; 
	    \end{axis}     
	\end{scope}

\end{tikzpicture}  
}

\newcommand{\distIII}{
\begin{tikzpicture}

	\begin{scope}[SCOPEPROPERTIESDISTRIBUTION] 
	    \begin{axis}[height=\AXISOPTIONHEIGHT, width=\AXISOPTIONWIDTH, grid=major, grid style={line width=.1pt, draw=gray!20}, xticklabel style={yshift=-1pt}, xlabel={random variable}, ylabel={probability density}, ylabel style={yshift=2pt}, xlabel style={yshift=-0.5pt}, xmin= 0.0, ymin=0.0, xmax = 1.0, ymax = 7.5]
          \addplot gnuplot[raw gnuplot, color=darkgreen, line width=1pt, no markers] {
            \GnuplotDefs
            plot [x=0:1] normaldistribution(x,0.5,0.1*0.1) with lines;
          };
          \addplot gnuplot[raw gnuplot, color=darkred, line width=1pt, dashed,  no markers] {
            \GnuplotDefs
            plot [x=0:1] normaldistribution(x,0.45,0.15*0.15) with lines;
          };
		
		\legend{{$\RANDVARONE \sim \mathcal{N}(0.5,0.1^2)$},{$\RANDVARTWO \sim \mathcal{N}(0.45,0.15^2)$}};
	    \end{axis}     
	\end{scope}
	
\end{tikzpicture}  
}

\newcommand{\distIV}{
\begin{tikzpicture}
	
	\begin{scope}[SCOPEPROPERTIESDISTRIBUTION] 
	    \begin{axis}[height=\AXISOPTIONHEIGHT, width=\AXISOPTIONWIDTH, grid=major, grid style={line width=.1pt, draw=gray!20}, xticklabel style={yshift=-1pt}, xlabel={random variable}, ylabel={probability density}, ylabel style={yshift=-4pt}, xlabel style={yshift=-0.5pt}, xmin= 0.0, ymin=0.0, xmax = 1.0, ymax = 11.5]

          \addplot gnuplot[raw gnuplot, color=darkgreen, line width=1pt, no markers] {
            \GnuplotDefs
            plot [x=0.001:0.999] beta(x,0.3,0.3) with lines;
          };
          \addplot gnuplot[raw gnuplot, color=darkred, line width=1pt, dashed,  no markers] {
            \GnuplotDefs
            plot [x=0:1] normaldistribution(x,0.3,0.05*0.05) with lines;
          };
		
		\legend{{$\RANDVARONE \sim \mathcal{B}(0.3,0.3)$},{$\RANDVARTWO \sim \mathcal{N}(0.3,0.05^2)$}}; 
	    \end{axis}     
	\end{scope}
 
\end{tikzpicture}
}

\newcommand{\distV}{
\begin{tikzpicture}

	\begin{scope}[SCOPEPROPERTIESDISTRIBUTION] 
	    \begin{axis}[height=\AXISOPTIONHEIGHT, width=\AXISOPTIONWIDTH, grid=major, grid style={line width=.1pt, draw=gray!20}, xticklabel style={yshift=-1pt}, xlabel={random variable}, ylabel={probability density}, ylabel style={yshift=-4pt}, xlabel style={yshift=-0.5pt}, xmin= 0.0, ymin=0.0, xmax = 1.0, ymax = 22.0]

		\addplot[mark=none, color=darkgreen, line width=1pt] coordinates { 
			(0.0,0.02)(0.3183099,0.02)(0.3183099,20.17367)(0.367879,20.17367)(0.367879,0.02)(1.0,0.02)
		}; 
		
		\addplot[mark=none, color=darkred, line width=1pt, dashed] coordinates { 
          (0.0,0.0)(0.25,0.0)(0.25,12.0)(0.33333,12.0)(0.33333,0.0)(1.0,0.0)
		};
		
        \legend{{$\RANDVARONE \sim \mathcal{U}(\frac{1}{\pi},\frac{1}{e})$},{$\RANDVARTWO \sim \mathcal{U}(\frac{1}{4},\frac{1}{3})$}}; 
	    \end{axis}     
	\end{scope}

\end{tikzpicture}  
}

\newcommand{\PWCOLOR}{blue}
\newcommand{\EmpCOLOR}{red}
\newcommand{\RedCOLOR}{black}
\newcommand{\HiACOLOR}{blue}
\newcommand{\HiBCOLOR}{red}
\newcommand{\HiCCOLOR}{black}
\newcommand{\UnACOLOR}{blue}
\newcommand{\UnBCOLOR}{red}
\newcommand{\GauCOLOR}{blue}
\newcommand{\PWLINESTYLE}{dashed}
\newcommand{\EmpLINESTYLE}{dashed}
\newcommand{\RedLINESTYLE}{dashed}
\newcommand{\HiALINESTYLE}{solid}
\newcommand{\HiBLINESTYLE}{solid}
\newcommand{\HiCLINESTYLE}{solid}
\newcommand{\UnALINESTYLE}{dotted}
\newcommand{\UnBLINESTYLE}{dotted}
\newcommand{\GauLINESTYLE}{dashdotted}

\newcommand{\compQualityI}{
\begin{tikzpicture}

	\begin{scope}[SCOPEPROPERTIESDISTRIBUTION]
      \begin{axis}[height=\AXISOPTIONHEIGHT, width=\AXISOPTIONWIDTH, xmode=log, ymode=log, grid=major, grid style={line width=.1pt, draw=gray!20}, xlabel={$N$}, ylabel={absolute error}, ylabel style={yshift=0}, xlabel style={yshift=0}, xmin= 1.0, xmax = 1000000.0, ymax = 1.4,  no markers]
        \addplot [color=\PWCOLOR , \PWLINESTYLE , line width=1pt] table[x index=0, y index=9, col sep=space]{./program/Distribution_errors/Distr_0_Pairwise_data.txt};
        \addplot [color=\EmpCOLOR, \EmpLINESTYLE, line width=1pt] table[x index=0, y index=9, col sep=space]{./program/Distribution_errors/Distr_0_Empirical_data.txt};
        \addplot [color=\RedCOLOR, \RedLINESTYLE, line width=1pt] table[x index=0, y index=9, col sep=space]{./program/Distribution_errors/Distr_0_Reduced_data.txt};
        \addplot [color=\HiACOLOR, \HiALINESTYLE, line width=1pt] table[x index=0, y index=9, col sep=space]{./program/Distribution_errors/Distr_0_Hist01_data.txt};
        \addplot [color=\HiBCOLOR, \HiBLINESTYLE, line width=1pt] table[x index=0, y index=9, col sep=space]{./program/Distribution_errors/Distr_0_Hist005_data.txt};
        \addplot [color=\HiCCOLOR, \HiCLINESTYLE, line width=1pt] table[x index=0, y index=9, col sep=space]{./program/Distribution_errors/Distr_0_Hist001_data.txt};
        \addplot [color=\UnACOLOR, \UnALINESTYLE, line width=1pt] table[x index=0, y index=9, col sep=space]{./program/Distribution_errors/Distr_0_Uniform1_data.txt};
        \addplot [color=\UnBCOLOR, \UnBLINESTYLE, line width=1pt] table[x index=0, y index=9, col sep=space]{./program/Distribution_errors/Distr_0_Uniform2_data.txt};
        \addplot [color=\GauCOLOR, \GauLINESTYLE, line width=1pt] table[x index=0, y index=9, col sep=space]{./program/Distribution_errors/Distr_0_Gauss_data.txt};
	    \end{axis}     
	\end{scope}

\end{tikzpicture}  
}
\newcommand{\compQualityII}{
\begin{tikzpicture}

	\begin{scope}[SCOPEPROPERTIESDISTRIBUTION]
      \begin{axis}[height=\AXISOPTIONHEIGHT, width=\AXISOPTIONWIDTH, xmode=log, ymode=log, grid=major, grid style={line width=.1pt, draw=gray!20}, xlabel={$N$}, ylabel={absolute error}, ylabel style={yshift=0}, xlabel style={yshift=0}, xmin= 1.0, xmax = 1000000.0, ymax = 1.0,  no markers]
        \addplot [color=\PWCOLOR , \PWLINESTYLE , line width=1pt] table[x index=0, y index=9, col sep=space]{./program/Distribution_errors/Distr_1_Pairwise_data.txt};
        \addplot [color=\EmpCOLOR, \EmpLINESTYLE, line width=1pt] table[x index=0, y index=9, col sep=space]{./program/Distribution_errors/Distr_1_Empirical_data.txt};
        \addplot [color=\RedCOLOR, \RedLINESTYLE, line width=1pt] table[x index=0, y index=9, col sep=space]{./program/Distribution_errors/Distr_1_Reduced_data.txt};
        \addplot [color=\HiACOLOR, \HiALINESTYLE, line width=1pt] table[x index=0, y index=9, col sep=space]{./program/Distribution_errors/Distr_1_Hist01_data.txt};
        \addplot [color=\HiBCOLOR, \HiBLINESTYLE, line width=1pt] table[x index=0, y index=9, col sep=space]{./program/Distribution_errors/Distr_1_Hist005_data.txt};
        \addplot [color=\HiCCOLOR, \HiCLINESTYLE, line width=1pt] table[x index=0, y index=9, col sep=space]{./program/Distribution_errors/Distr_1_Hist001_data.txt};
        \addplot [color=\UnACOLOR, \UnALINESTYLE, line width=1pt] table[x index=0, y index=9, col sep=space]{./program/Distribution_errors/Distr_1_Uniform1_data.txt};
        \addplot [color=\UnBCOLOR, \UnBLINESTYLE, line width=1pt] table[x index=0, y index=9, col sep=space]{./program/Distribution_errors/Distr_1_Uniform2_data.txt};
        \addplot [color=\GauCOLOR, \GauLINESTYLE, line width=1pt] table[x index=0, y index=9, col sep=space]{./program/Distribution_errors/Distr_1_Gauss_data.txt};
	    \end{axis}     
	\end{scope}

\end{tikzpicture}  
}
\newcommand{\compQualityIII}{
\begin{tikzpicture}

	\begin{scope}[SCOPEPROPERTIESDISTRIBUTION]
      \begin{axis}[height=\AXISOPTIONHEIGHT, width=\AXISOPTIONWIDTH, xmode=log, ymode=log, grid=major, grid style={line width=.1pt, draw=gray!20}, xlabel={$N$}, ylabel={absolute error}, ylabel style={yshift=0}, xlabel style={yshift=0}, xmin= 1.0, xmax = 1000000.0, ymax = 1.0,  no markers]
        \addplot [color=\PWCOLOR , \PWLINESTYLE , line width=1pt] table[x index=0, y index=9, col sep=space]{./program/Distribution_errors/Distr_2_Pairwise_data.txt};
        \addplot [color=\EmpCOLOR, \EmpLINESTYLE, line width=1pt] table[x index=0, y index=9, col sep=space]{./program/Distribution_errors/Distr_2_Empirical_data.txt};
        \addplot [color=\RedCOLOR, \RedLINESTYLE, line width=1pt] table[x index=0, y index=9, col sep=space]{./program/Distribution_errors/Distr_2_Reduced_data.txt};
        \addplot [color=\HiACOLOR, \HiALINESTYLE, line width=1pt] table[x index=0, y index=9, col sep=space]{./program/Distribution_errors/Distr_2_Hist01_data.txt};
        \addplot [color=\HiBCOLOR, \HiBLINESTYLE, line width=1pt] table[x index=0, y index=9, col sep=space]{./program/Distribution_errors/Distr_2_Hist005_data.txt};
        \addplot [color=\HiCCOLOR, \HiCLINESTYLE, line width=1pt] table[x index=0, y index=9, col sep=space]{./program/Distribution_errors/Distr_2_Hist001_data.txt};
        \addplot [color=\UnACOLOR, \UnALINESTYLE, line width=1pt] table[x index=0, y index=9, col sep=space]{./program/Distribution_errors/Distr_2_Uniform1_data.txt};
        \addplot [color=\UnBCOLOR, \UnBLINESTYLE, line width=1pt] table[x index=0, y index=9, col sep=space]{./program/Distribution_errors/Distr_2_Uniform2_data.txt};
        \addplot [color=\GauCOLOR, \GauLINESTYLE, line width=1pt] table[x index=0, y index=9, col sep=space]{./program/Distribution_errors/Distr_2_Gauss_data.txt};
	    \end{axis}     
	\end{scope}

\end{tikzpicture}  
}
\newcommand{\compQualityIV}{
\begin{tikzpicture}

	\begin{scope}[SCOPEPROPERTIESDISTRIBUTION]
      \begin{axis}[height=\AXISOPTIONHEIGHT, width=\AXISOPTIONWIDTH, xmode=log, ymode=log, grid=major, grid style={line width=.1pt, draw=gray!20}, xlabel={$N$}, ylabel={absolute error}, ylabel style={yshift=0}, xlabel style={yshift=0}, xmin= 1.0, xmax = 1000000.0, ymax = 1.0,  no markers]
        \addplot [color=\PWCOLOR , \PWLINESTYLE , line width=1pt] table[x index=0, y index=9, col sep=space]{./program/Distribution_errors/Distr_3_Pairwise_data.txt};
        \addplot [color=\EmpCOLOR, \EmpLINESTYLE, line width=1pt] table[x index=0, y index=9, col sep=space]{./program/Distribution_errors/Distr_3_Empirical_data.txt};
        \addplot [color=\RedCOLOR, \RedLINESTYLE, line width=1pt] table[x index=0, y index=9, col sep=space]{./program/Distribution_errors/Distr_3_Reduced_data.txt};
        \addplot [color=\HiACOLOR, \HiALINESTYLE, line width=1pt] table[x index=0, y index=9, col sep=space]{./program/Distribution_errors/Distr_3_Hist01_data.txt};
        \addplot [color=\HiBCOLOR, \HiBLINESTYLE, line width=1pt] table[x index=0, y index=9, col sep=space]{./program/Distribution_errors/Distr_3_Hist005_data.txt};
        \addplot [color=\HiCCOLOR, \HiCLINESTYLE, line width=1pt] table[x index=0, y index=9, col sep=space]{./program/Distribution_errors/Distr_3_Hist001_data.txt};
        \addplot [color=\UnACOLOR, \UnALINESTYLE, line width=1pt] table[x index=0, y index=9, col sep=space]{./program/Distribution_errors/Distr_3_Uniform1_data.txt};
        \addplot [color=\UnBCOLOR, \UnBLINESTYLE, line width=1pt] table[x index=0, y index=9, col sep=space]{./program/Distribution_errors/Distr_3_Uniform2_data.txt};
        \addplot [color=\GauCOLOR, \GauLINESTYLE, line width=1pt] table[x index=0, y index=9, col sep=space]{./program/Distribution_errors/Distr_3_Gauss_data.txt};
	    \end{axis}     
	\end{scope}

\end{tikzpicture}  
}
\newcommand{\compQualityV}{
\begin{tikzpicture}

	\begin{scope}[SCOPEPROPERTIESDISTRIBUTION]
      \begin{axis}[height=\AXISOPTIONHEIGHT, width=\AXISOPTIONWIDTH, xmode=log, ymode=log, grid=major, grid style={line width=.1pt, draw=gray!20}, xlabel={$N$}, ylabel={absolute error}, ylabel style={yshift=0}, xlabel style={yshift=0}, xmin= 1.0, xmax = 1000000.0, ymax = 1.0,  no markers]
        \addplot [color=\PWCOLOR , \PWLINESTYLE , line width=1pt] table[x index=0, y index=9, col sep=space]{./program/Distribution_errors/Distr_4_Pairwise_data.txt};
        \addplot [color=\EmpCOLOR, \EmpLINESTYLE, line width=1pt] table[x index=0, y index=9, col sep=space]{./program/Distribution_errors/Distr_4_Empirical_data.txt};
        \addplot [color=\RedCOLOR, \RedLINESTYLE, line width=1pt] table[x index=0, y index=9, col sep=space]{./program/Distribution_errors/Distr_4_Reduced_data.txt};
        \addplot [color=\HiACOLOR, \HiALINESTYLE, line width=1pt] table[x index=0, y index=9, col sep=space]{./program/Distribution_errors/Distr_4_Hist01_data.txt};
        \addplot [color=\HiBCOLOR, \HiBLINESTYLE, line width=1pt] table[x index=0, y index=9, col sep=space]{./program/Distribution_errors/Distr_4_Hist005_data.txt};
        \addplot [color=\HiCCOLOR, \HiCLINESTYLE, line width=1pt] table[x index=0, y index=9, col sep=space]{./program/Distribution_errors/Distr_4_Hist001_data.txt};
        \addplot [color=\UnACOLOR, \UnALINESTYLE, line width=1pt] table[x index=0, y index=9, col sep=space]{./program/Distribution_errors/Distr_4_Uniform1_data.txt};
        \addplot [color=\UnBCOLOR, \UnBLINESTYLE, line width=1pt] table[x index=0, y index=9, col sep=space]{./program/Distribution_errors/Distr_4_Uniform2_data.txt};
        \addplot [color=\GauCOLOR, \GauLINESTYLE, line width=1pt] table[x index=0, y index=9, col sep=space]{./program/Distribution_errors/Distr_4_Gauss_data.txt};
	    \end{axis}     
	\end{scope}

\end{tikzpicture}  
}

\newcommand{\compQualityLegendI}{
\begin{tikzpicture}
	\begin{scope}[SCOPEPROPERTIESDISTRIBUTION]
      \begin{axis}[width=2cm, hide axis, xmin=1.0, ymin=0.0, xmax = 1000000.0, ymax = 1.0, legend style={draw=white, legend cell align=left, row sep=1.5pt}]
        \addlegendimage{color=\UnACOLOR, \UnALINESTYLE, line width=1pt}; \addlegendentry{$\AUniform^1$};
        \addlegendimage{color=\UnBCOLOR, \UnBLINESTYLE, line width=1pt}; \addlegendentry{$\AUniform^2$};
        \addlegendimage{color=\HiACOLOR, \HiALINESTYLE, line width=1pt}; \addlegendentry{$\AHistogram{0.1}$};
        \addlegendimage{color=\GauCOLOR, \GauLINESTYLE, line width=1pt}; \addlegendentry{$\AGauss$};
        \addlegendimage{color=\HiBCOLOR, \HiBLINESTYLE, line width=1pt}; \addlegendentry{$\AHistogram{0.05}$};
        \addlegendimage{color=\PWCOLOR , \PWLINESTYLE , line width=1pt}; \addlegendentry{$\APW$};
        \addlegendimage{color=\HiCCOLOR, \HiCLINESTYLE, line width=1pt}; \addlegendentry{$\AHistogram{0.01}$};
        \addlegendimage{color=\RedCOLOR, \RedLINESTYLE, line width=1pt}; \addlegendentry{$\AReduced$};
        \addlegendimage{color=\EmpCOLOR, \EmpLINESTYLE, line width=1pt}; \addlegendentry{$\AEmpirical$};
        \addlegendimage{color=white}; \addlegendentry{\phantom{A}};
        \addlegendimage{color=white}; \addlegendentry{\phantom{A}};
	    \end{axis}     
	\end{scope}
\end{tikzpicture}  
}
\newcommand{\compQualityLegendII}{
\begin{tikzpicture}
	\begin{scope}[SCOPEPROPERTIESDISTRIBUTION]
      \begin{axis}[width=2cm, hide axis, xmin=1.0, ymin=0.0, xmax = 1000000.0, ymax = 1.0, legend style={draw=white, legend cell align=left, row sep=1.5pt}]
        \addlegendimage{color=\GauCOLOR, \GauLINESTYLE, line width=1pt}; \addlegendentry{$\AGauss$};
        \addlegendimage{color=\PWCOLOR , \PWLINESTYLE , line width=1pt}; \addlegendentry{$\APW$};
        \addlegendimage{color=\HiCCOLOR, \HiCLINESTYLE, line width=1pt}; \addlegendentry{$\AHistogram{0.01}$};
        \addlegendimage{color=\RedCOLOR, \RedLINESTYLE, line width=1pt}; \addlegendentry{$\AReduced$};
        \addlegendimage{color=\EmpCOLOR, \EmpLINESTYLE, line width=1pt}; \addlegendentry{$\AEmpirical$};
        \addlegendimage{color=\HiBCOLOR, \HiBLINESTYLE, line width=1pt}; \addlegendentry{$\AHistogram{0.05}$};
        \addlegendimage{color=\HiACOLOR, \HiALINESTYLE, line width=1pt}; \addlegendentry{$\AHistogram{0.1}$};
        \addlegendimage{color=\UnBCOLOR, \UnBLINESTYLE, line width=1pt}; \addlegendentry{$\AUniform^2$};
        \addlegendimage{color=\UnACOLOR, \UnALINESTYLE, line width=1pt}; \addlegendentry{$\AUniform^1$};
        \addlegendimage{color=white}; \addlegendentry{\phantom{A}};
        \addlegendimage{color=white}; \addlegendentry{\phantom{A}};
	    \end{axis}     
	\end{scope}
\end{tikzpicture}  
}
\newcommand{\compQualityLegendIII}{
\begin{tikzpicture}
	\begin{scope}[SCOPEPROPERTIESDISTRIBUTION]
      \begin{axis}[width=2cm, hide axis, xmin=1.0, ymin=0.0, xmax = 1000000.0, ymax = 1.0, legend style={draw=white, legend cell align=left, row sep=1.5pt}]
        \addlegendimage{color=\UnACOLOR, \UnALINESTYLE, line width=1pt}; \addlegendentry{$\AUniform^1$};
        \addlegendimage{color=\UnBCOLOR, \UnBLINESTYLE, line width=1pt}; \addlegendentry{$\AUniform^2$};
        \addlegendimage{color=\HiACOLOR, \HiALINESTYLE, line width=1pt}; \addlegendentry{$\AHistogram{0.1}$};
        \addlegendimage{color=\HiBCOLOR, \HiBLINESTYLE, line width=1pt}; \addlegendentry{$\AHistogram{0.05}$};
        \addlegendimage{color=\PWCOLOR , \PWLINESTYLE , line width=1pt}; \addlegendentry{$\APW$};
        \addlegendimage{color=\RedCOLOR, \RedLINESTYLE, line width=1pt}; \addlegendentry{$\AReduced$};
        \addlegendimage{color=\EmpCOLOR, \EmpLINESTYLE, line width=1pt}; \addlegendentry{$\AEmpirical$};
        \addlegendimage{color=\HiCCOLOR, \HiCLINESTYLE, line width=1pt}; \addlegendentry{$\AHistogram{0.01}$};
        \addlegendimage{color=\GauCOLOR, \GauLINESTYLE, line width=1pt}; \addlegendentry{$\AGauss$};
        \addlegendimage{color=white}; \addlegendentry{\phantom{A}};
        \addlegendimage{color=white}; \addlegendentry{\phantom{A}};
	    \end{axis}     
	\end{scope}
\end{tikzpicture}  
}
\newcommand{\compQualityLegendIV}{
\begin{tikzpicture}
	\begin{scope}[SCOPEPROPERTIESDISTRIBUTION]
      \begin{axis}[width=2cm, hide axis, xmin=1.0, ymin=0.0, xmax = 1000000.0, ymax = 1.0, legend style={draw=white, legend cell align=left, row sep=1.5pt}]
        \addlegendimage{color=\UnACOLOR, \UnALINESTYLE, line width=1pt}; \addlegendentry{$\AUniform^1$};
        \addlegendimage{color=\GauCOLOR, \GauLINESTYLE, line width=1pt}; \addlegendentry{$\AGauss$};
        \addlegendimage{color=\UnBCOLOR, \UnBLINESTYLE, line width=1pt}; \addlegendentry{$\AUniform^2$};
        \addlegendimage{color=\HiACOLOR, \HiALINESTYLE, line width=1pt}; \addlegendentry{$\AHistogram{0.1}$};
        \addlegendimage{color=\HiBCOLOR, \HiBLINESTYLE, line width=1pt}; \addlegendentry{$\AHistogram{0.05}$};
        \addlegendimage{color=\PWCOLOR , \PWLINESTYLE , line width=1pt}; \addlegendentry{$\APW$};
        \addlegendimage{color=\RedCOLOR, \RedLINESTYLE, line width=1pt}; \addlegendentry{$\AReduced$};
        \addlegendimage{color=\EmpCOLOR, \EmpLINESTYLE, line width=1pt}; \addlegendentry{$\AEmpirical$};
        \addlegendimage{color=\HiCCOLOR, \HiCLINESTYLE, line width=1pt}; \addlegendentry{$\AHistogram{0.01}$};
        \addlegendimage{color=white}; \addlegendentry{\phantom{A}};
        \addlegendimage{color=white}; \addlegendentry{\phantom{A}};
	    \end{axis}     
	\end{scope}
\end{tikzpicture}  
}
\newcommand{\compQualityLegendV}{
\begin{tikzpicture}
	\begin{scope}[SCOPEPROPERTIESDISTRIBUTION]
      \begin{axis}[width=2cm, hide axis, xmin=1.0, ymin=0.0, xmax = 1000000.0, ymax = 1.0, legend style={draw=white, legend cell align=left, row sep=1.5pt}]
        \addlegendimage{color=\HiACOLOR, \HiALINESTYLE, line width=1pt}; \addlegendentry{$\AHistogram{0.1}$};
        \addlegendimage{color=\HiBCOLOR, \HiBLINESTYLE, line width=1pt}; \addlegendentry{$\AHistogram{0.05}$};
        \addlegendimage{color=\GauCOLOR, \GauLINESTYLE, line width=1pt}; \addlegendentry{$\AGauss$};
        \addlegendimage{color=\HiCCOLOR, \HiCLINESTYLE, line width=1pt}; \addlegendentry{$\AHistogram{0.01}$};
        \addlegendimage{color=\PWCOLOR , \PWLINESTYLE , line width=1pt}; \addlegendentry{$\APW$};
        \addlegendimage{color=\UnBCOLOR, \UnBLINESTYLE, line width=1pt}; \addlegendentry{$\AUniform^2$};
        \addlegendimage{color=\RedCOLOR, \RedLINESTYLE, line width=1pt}; \addlegendentry{$\AReduced$};
        \addlegendimage{color=\EmpCOLOR, \EmpLINESTYLE, line width=1pt}; \addlegendentry{$\AEmpirical$};
        \addlegendimage{color=\UnACOLOR, \UnALINESTYLE, line width=1pt}; \addlegendentry{$\AUniform^1$};
        \addlegendimage{color=white}; \addlegendentry{\phantom{A}};
        \addlegendimage{color=white}; \addlegendentry{\phantom{A}};
	    \end{axis}     
	\end{scope}
\end{tikzpicture}  
}

\tikzstyle{pareto_hist} = [only marks, mark=o, mark options={scale=0.5, line width=0.2pt}, cyan!80!blue, opacity=0.7]
\tikzstyle{pareto_threestage} = [only marks, mark=triangle, mark options={scale=0.6, line width=0.2pt}, midblue!70!magenta, opacity=0.7]
\tikzstyle{pareto_reduced} = [only marks, mark=x, mark options={scale=0.8}, azure, draw opacity=1.0, fill opacity=0.7]
\tikzstyle{pareto_naive} = [only marks, mark=x, mark options={scale=0.6, line width=0.2pt}, red, draw opacity=0.5]
\tikzstyle{pareto_200} = [only marks, mark=square, mark options={scale=1.0}, green, draw opacity=1.0]

\newcommand{\frontsUZDTone}[3]{
\begin{tikzpicture}[font=\small]
  \begin{axis}[height=0.60\textwidth, width=\textwidth, grid=major, xlabel={$f_1$}, ylabel={$f_2$}, ylabel style={yshift=-0}, xlabel style={yshift=0}, enlarge x limits = 0.04, enlarge y limits = 0.04, 
  xmin=0.0, xmax=1.0, ymin=0.0, ymax=1.25, xtick=\empty, ytick=\empty, title=UZDT1 - #3, title style = {yshift=-8}]   
    \addlegendimage{no markers, red, draw opacity=0.6, line width=0.8pt}
    \addlegendentry{ref.}
    
    \addlegendimage{pareto_200}
    \addlegendentry{200G}
    
    \addlegendimage{pareto_reduced}
    \addlegendentry{400G}
    
    \addplot [only marks, red, draw opacity=0.6, mark size=0.1pt, smooth] table[x index=3, y index=4]{./results/fronts/means/uzdt1/reference};    
    \addplot [pareto_200] table[x index=3, y index=4]{./results/fronts/means/uzdt1/200/#1.#2.tsv};    
    \addplot [pareto_reduced] table[x index=3, y index=4, smooth]{./results/fronts/means/uzdt1/400/#1.#2.tsv};
  \end{axis}  
\end{tikzpicture}  
}

\newcommand{\frontsUZDTthree}[3]{
\begin{tikzpicture}[font=\small]
  \begin{axis}[height=0.60\textwidth, width=\textwidth, grid=major, xlabel={$f_1$}, ylabel={$f_2$}, ylabel style={yshift=-0}, xlabel style={yshift=0}, enlarge x limits = 0.04, enlarge y limits = 0.04, 
  xmin=0.0, xmax=1.0, ymin=-0.85, ymax=1.45, xtick=\empty, ytick=\empty, title=UZDT3 - #3, title style = {yshift=-8}] 
    \addlegendimage{no markers, red, draw opacity=0.6, line width=0.8pt}
    \addlegendentry{ref.}
    
    \addlegendimage{pareto_200}
    \addlegendentry{200G}
    
    \addlegendimage{pareto_reduced}
    \addlegendentry{400G}
    
    \addplot [only marks, red, draw opacity=0.6, mark size=0.1pt, smooth] table[x index=3, y index=4]{./results/fronts/means/uzdt3/reference};    
    \addplot [pareto_200] table[x index=3, y index=4]{./results/fronts/means/uzdt3/200/#1.#2.tsv};    
    \addplot [pareto_reduced] table[x index=3, y index=4]{./results/fronts/means/uzdt3/400/#1.#2.tsv};
    
    \draw[very thick, white] (axis cs:0.075, 0.67) -- (axis cs:0.185, 0.67);
    \draw[very thick, white] (axis cs:0.24, 0.242) -- (axis cs:0.42, 0.242);
    \draw[very thick, white] (axis cs:0.44, -0.125) -- (axis cs:0.64, -0.125);
    \draw[very thick, white] (axis cs:0.64, -0.458) -- (axis cs:0.845, -0.458);

  \end{axis}  
\end{tikzpicture}  
}

\tikzstyle{eps_hist} = [midgreen, opacity=0.5]
\tikzstyle{eps_reduced} = [lightblue!80!magenta]
\tikzstyle{eps_stage} = [red, densely dotted]
\tikzstyle{eps_naive} = [densely dashed, opacity=0.5]

\tikzstyle{div_hist} = [midgreen!70!darkgreen, densely dotted]
\tikzstyle{div_reduced} = [midblue!77!magenta, densely dashed]
\tikzstyle{div_stage} = [red, opacity=0.3]
\tikzstyle{div_naive} = [opacity=0.3]

\newcommand{\addlegends}{
  \begin{tikzpicture}[font=\scriptsize]
    \begin{axis}[height=0.165\textwidth, width=1.0\textwidth, ticks=none, legend style={draw=none, legend cell align=left, row sep = 2.0pt}, legend columns=4, xmin=0, xmax=1, ymin=0, ymax=1]
        \addlegendimage{eps_reduced}
        \addlegendentry{$\AReduced$ ($\epsilon$--dom.)$\qquad$};
        \addlegendimage{eps_hist}
        \addlegendentry{$\AHistogram{0.01}$ ($\epsilon$-dom.)$\qquad$};
        \addlegendimage{eps_stage}
        \addlegendentry{$\AThreeStage$ ($\epsilon$--dom.)$\qquad$};
        \addlegendimage{eps_naive}
        \addlegendentry{$\AMeanBased$ ($\epsilon$--dom.)};
    
        \addlegendimage{div_reduced}
        \addlegendentry{$\AReduced$ (DCI)};
        \addlegendimage{div_hist}
        \addlegendentry{$\AHistogram{0.01}$ (DCI)};
        \addlegendimage{div_stage}
        \addlegendentry{$\AThreeStage$ (DCI)};
        \addlegendimage{div_naive}
        \addlegendentry{$\AMeanBased$ (DCI)};
    \end{axis}
  \end{tikzpicture} 
}

\newcommand{\dominance}[1]{
    \addplot [eps_hist] table[x index=0, y index=1]{./results/dominance/#1/hist.stats};
    \addplot [eps_reduced] table[x index=0, y index=1]{./results/dominance/#1/reduced.stats};
    \addplot [eps_stage] table[x index=0, y index=1]{./results/dominance/#1/threestage.stats};
    \addplot [eps_naive] table[x index=0, y index=1]{./results/dominance/#1/naive.stats}; 
}

\newcommand{\diversity}[1]{
    \addplot [div_hist] table[x index=0, y index=1]{./results/diversity/#1/hist.tsv};
    \addplot [div_reduced] table[x index=0, y index=1]{./results/diversity/#1/reduced.tsv};
    \addplot [div_stage] table[x index=0, y index=1]{./results/diversity/#1/threestage.tsv};
    \addplot [div_naive] table[x index=0, y index=1]{./results/diversity/#1/naive.tsv};
}

\newcommand{\dominanceDiversityUZDTone}{
\resizebox{\textwidth}{!}{
\begin{tikzpicture}[font=\scriptsize]
  \begin{axis}[height=0.95\textwidth, width=\textwidth, xlabel style={yshift=2pt}, axis y line*=left, xlabel={generation}, grid=major, ylabel={$\epsilon$-dominance}, enlargelimits = 0.02, yticklabel style={/pgf/number format/fixed}, ymin=0, ymax = 0.5]
    \dominance{uzdt1}
  \end{axis}
  
  \begin{axis}[height=0.95\textwidth, width=\textwidth, xlabel style={yshift=2pt}, axis y line*=right, axis x line=none, grid=none, ylabel={diversity comparison indicator (DCI)}, enlargelimits = 0.02, yticklabel style={/pgf/number format/fixed}, ymin=0, ymax = 1] 
    \diversity{uzdt1}
  \end{axis} 
\end{tikzpicture}  
}
}

\newcommand{\dominanceDiversityUZDTthree}{
\resizebox{\textwidth}{!}{
\begin{tikzpicture}[font=\scriptsize]
  \begin{axis}[height=0.95\textwidth, width=\textwidth, xlabel style={yshift=2pt}, axis y line*=left, xlabel={generation}, grid=major, ylabel={$\epsilon$-dominance}, enlargelimits = 0.02, yticklabel style={/pgf/number format/fixed}, ymin=0, ymax = 0.5]
    \dominance{uzdt3}
  \end{axis}
  
  \begin{axis}[height=0.95\textwidth, width=\textwidth, xlabel style={yshift=2pt}, axis y line*=right, axis x line=none, grid=none, ylabel={diversity comparison indicator (DCI)}, enlargelimits = 0.02, yticklabel style={/pgf/number format/fixed}, ymin=0, ymax = 1] 
    \diversity{uzdt3}
  \end{axis} 
\end{tikzpicture}  
}
}

\newcommand{\dominanceDiversityUDTLZtwo}{
\resizebox{\textwidth}{!}{
\begin{tikzpicture}[font=\scriptsize]
  \begin{axis}[height=0.95\textwidth, width=\textwidth, xlabel style={yshift=2pt}, axis y line*=left, xlabel={generation}, grid=major, ylabel={$\epsilon$-dominance}, enlargelimits = 0.02, yticklabel style={/pgf/number format/fixed}, ymin=0, ymax = 0.4]
    \dominance{udtlz2}
  \end{axis}
  
  \begin{axis}[height=0.95\textwidth, width=\textwidth, xlabel style={yshift=2pt}, axis y line*=right, axis x line=none, grid=none, ylabel={diversity comparison indicator (DCI)}, enlargelimits = 0.02, yticklabel style={/pgf/number format/fixed}, ymin=0.1, ymax = 0.9, ytick={0.1,0.3,0.5,0.7,0.9}] 
    \diversity{udtlz2}
  \end{axis} 
\end{tikzpicture}  
}
}

\newcommand{\dominanceDiversityUDTLZseven}{
\resizebox{\textwidth}{!}{
\begin{tikzpicture}[font=\scriptsize]
  \begin{axis}[height=0.95\textwidth, width=\textwidth, xlabel style={yshift=2pt}, axis y line*=left, xlabel={generation}, grid=major, ylabel={$\epsilon$-dominance}, enlargelimits = 0.02, yticklabel style={/pgf/number format/fixed}, ymin=0.03, ymax = 0.28, ytick={0.05,0.1,0.15,0.20,0.25}]
    \dominance{udtlz7}
  \end{axis}
  
  \begin{axis}[height=0.95\textwidth, width=\textwidth, xlabel style={yshift=2pt}, axis y line*=right, axis x line=none, grid=none, ylabel={diversity comparison indicator (DCI)}, enlargelimits = 0.02, yticklabel style={/pgf/number format/fixed}, ymin=0.24, ymax = 0.99, ytick={0.3,0.45,0.6,0.75,0.9}]  
    \diversity{udtlz7}
  \end{axis} 
\end{tikzpicture}  
}
}

\newcommand{\distances}{
  \begin{tikzpicture}[font=\footnotesize]
    \begin{axis}[
		  width=1.015\textwidth, 
		  height=7cm, 
		  ybar,
		  bar width=5pt,
		  grid style={line width=.1pt,draw=gray!10}, grid=major, 
		  enlargelimits=0.05,
		  ymin = 0.0012, 
		  ymax = 0.25,
		  xtick align=inside,
		  ylabel={average diagonal distance},
		  ylabel style={yshift=-4},
		  xlabel style={yshift=20},
		  legend cell align=left, 
		  legend columns=5,
		  legend pos=north west,
		  xtick={1,2,3,4,5,6,7,8,9,10,11,12}, 
		  xticklabels={UZDT1, UZDT2, UZDT3, UZDT4, UZDT5, UZDT6, UDTLZ1, UDTLZ2, UDTLZ3, UDTLZ4, UDTLZ5, UDTLZ6},
		  xticklabel style={font=\scriptsize},
		  ymode = log, 
		  log origin = infty
      ]

      \addplot[cyan!70!blue, fill=cyan!30!lightblue] 	coordinates {(1,0.0342)(2,0.0389)(3,0.0368)(4,0.0053)(5,0.0124)(6,0.0072)(7,0.0484)(8,0.0413)(9,0.0396)(10,0.0414)(11,0.0404)(12,0.1812)};
      \addplot[darkgreen!80, fill=darkgreen!40 ] 		coordinates {(1,0.0342)(2,0.0403)(3,0.0305)(4,0.0051)(5,0.0124)(6,0.0072)(7,0.0412)(8,0.0413)(9,0.0396)(10,0.0413)(11,0.0405)(12,0.1846)};
      \addplot[darkbrown, fill=lightbrown]              coordinates {(1,0.0368)(2,0.0388)(3,0.0380)(4,0.0051)(5,0.0087)(6,0.0070)(7,0.0467)(8,0.0396)(9,0.0337)(10,0.0390)(11,0.0392)(12,0.1551)}; 
      \addplot[darkred, fill=darkred!50] 	            coordinates {(1,0.0396)(2,0.0363)(3,0.0389)(4,0.0045)(5,0.0127)(6,0.0064)(7,0.0417)(8,0.0405)(9,0.0388)(10,0.0411)(11,0.0445)(12,0.1532)};
      \legend{$\AReduced$,$\AHistogram{0.01}$~\cite{KBT18},$\AThreeStage$~\cite{KMGT15},$\AMeanBased$~\cite{MAG12}}
      
    \end{axis}
  \end{tikzpicture}
}

\newcommand{\cdfExample}{
\begin{tikzpicture}
  \begin{scope}[scale=1.0, font=\small]
    \begin{axis}[grid style={line width=.1pt, draw=gray!20},major grid style={line width=.1pt,draw=gray!20}, grid=major, yticklabel style={/pgf/number format/fixed}, 
      width=8.4cm, height=6.6cm, legend pos = north west,  xlabel={objective value}, xtick={0.48,0.903,1.92,2.2}, xticklabels={$\underline{\obj(A)}$,$\underline{\obj(B)}$,$\overline{\obj(A)}$,$\overline{\obj(B)}$}, ylabel={probability}, ytick={0.0,0.2,0.4,0.6,0.8,1.0}, ylabel style={yshift=0pt}, xlabel style={yshift=0pt}]
	
      \addplot [darkgreen, smooth, line width=0.6pt] coordinates {
        (0.48,0.0026)(0.537,0.0049)(0.572,0.0072)(0.598,0.0094)(0.618,0.0116)(0.636,0.0138)(0.651,0.0160)(0.664,0.0180)(0.676,0.0202)(0.687,0.0223)(0.696,0.0242)(0.706,0.0265)(0.714,0.0284)(0.722,0.0305)(0.73,0.0327)(0.737,0.0347)(0.744,0.0368)(0.75,0.0387)(0.756,0.0407)(0.762,0.0428)(0.768,0.0449)(0.773,0.0468)(0.779,0.0491)(0.784,0.0511)(0.789,0.0532)(0.794,0.0553)(0.798,0.0571)(0.803,0.0594)(0.807,0.0612)(0.811,0.0631)(0.815,0.0651)(0.819,0.0671)(0.823,0.0692)(0.827,0.0713)(0.831,0.0734)(0.835,0.0756)(0.838,0.0773)(0.842,0.0796)(0.845,0.0814)(0.849,0.0838)(0.852,0.0856)(0.855,0.0874)(0.859,0.0900)(0.862,0.0919)(0.865,0.0938)(0.868,0.0958)(0.871,0.0978)(0.874,0.0999)(0.877,0.1020)(0.88,0.1041)(0.882,0.1055)(0.885,0.1077)(0.888,0.1099)(0.891,0.1121)(0.893,0.1136)(0.896,0.1159)(0.899,0.1183)(0.901,0.1198)(0.904,0.1222)(0.906,0.1238)(0.909,0.1263)(0.911,0.1279)(0.914,0.1304)(0.916,0.1321)(0.918,0.1338)(0.921,0.1363)(0.923,0.1381)(0.925,0.1398)(0.928,0.1425)(0.93,0.1443)(0.932,0.1461)(0.934,0.1479)(0.937,0.1506)(0.939,0.1525)(0.941,0.1544)(0.943,0.1563)(0.945,0.1582)(0.947,0.1601)(0.949,0.1620)(0.951,0.1640)(0.953,0.1659)(0.955,0.1679)(0.957,0.1699)(0.959,0.1719)(0.961,0.1740)(0.963,0.1760)(0.965,0.1781)(0.967,0.1802)(0.969,0.1823)(0.971,0.1844)(0.973,0.1865)(0.975,0.1886)(0.977,0.1908)(0.979,0.1929)(0.981,0.1951)(0.982,0.1962)(0.984,0.1984)(0.986,0.2006)(0.988,0.2029)(0.99,0.2051)(0.991,0.2062)(0.993,0.2085)(0.995,0.2108)(0.997,0.2131)(0.998,0.2143)(1,0.2166)(1.002,0.2189)(1.004,0.2213)(1.005,0.2225)(1.007,0.2248)(1.009,0.2272)(1.01,0.2284)(1.012,0.2308)(1.014,0.2333)(1.015,0.2345)(1.017,0.2369)(1.019,0.2394)(1.02,0.2406)(1.022,0.2431)(1.023,0.2444)(1.025,0.2469)(1.027,0.2494)(1.028,0.2507)(1.03,0.2532)(1.031,0.2545)(1.033,0.2570)(1.035,0.2596)(1.036,0.2609)(1.038,0.2635)(1.039,0.2648)(1.041,0.2674)(1.042,0.2687)(1.044,0.2714)(1.045,0.2727)(1.047,0.2753)(1.048,0.2767)(1.05,0.2793)(1.051,0.2807)(1.053,0.2834)(1.054,0.2847)(1.056,0.2874)(1.057,0.2888)(1.059,0.2915)(1.06,0.2929)(1.062,0.2956)(1.063,0.2970)(1.065,0.2998)(1.066,0.3012)(1.067,0.3026)(1.069,0.3054)(1.07,0.3068)(1.072,0.3096)(1.073,0.3110)(1.075,0.3138)(1.076,0.3152)(1.077,0.3166)(1.079,0.3195)(1.08,0.3209)(1.082,0.3238)(1.083,0.3252)(1.084,0.3266)(1.086,0.3295)(1.087,0.3310)(1.089,0.3339)(1.09,0.3353)(1.091,0.3368)(1.093,0.3397)(1.094,0.3412)(1.096,0.3441)(1.097,0.3456)(1.098,0.3471)(1.1,0.3500)(1.101,0.3515)(1.102,0.3530)(1.104,0.3559)(1.105,0.3574)(1.106,0.3589)(1.108,0.3619)(1.109,0.3634)(1.11,0.3649)(1.112,0.3679)(1.113,0.3694)(1.114,0.3709)(1.116,0.3740)(1.117,0.3755)(1.118,0.3770)(1.12,0.3800)(1.121,0.3816)(1.122,0.3831)(1.124,0.3862)(1.125,0.3877)(1.126,0.3892)(1.128,0.3923)(1.129,0.3938)(1.13,0.3954)(1.131,0.3969)(1.133,0.4000)(1.134,0.4015)(1.135,0.4031)(1.137,0.4062)(1.138,0.4077)(1.139,0.4093)(1.141,0.4124)(1.142,0.4140)(1.143,0.4155)(1.144,0.4171)(1.146,0.4202)(1.147,0.4218)(1.148,0.4234)(1.15,0.4265)(1.151,0.4281)(1.152,0.4296)(1.153,0.4312)(1.155,0.4344)(1.156,0.4359)(1.157,0.4375)(1.158,0.4391)(1.16,0.4423)(1.161,0.4438)(1.162,0.4454)(1.164,0.4486)(1.165,0.4502)(1.166,0.4518)(1.167,0.4533)(1.169,0.4565)(1.17,0.4581)(1.171,0.4597)(1.172,0.4613)(1.174,0.4645)(1.175,0.4661)(1.176,0.4677)(1.177,0.4693)(1.179,0.4725)(1.18,0.4741)(1.181,0.4756)(1.182,0.4772)(1.184,0.4804)(1.185,0.4820)(1.186,0.4836)(1.187,0.4852)(1.189,0.4884)(1.19,0.4900)(1.191,0.4916)(1.192,0.4932)(1.194,0.4964)(1.195,0.4980)(1.196,0.4996)(1.197,0.5012)(1.199,0.5044)(1.2,0.5060)(1.201,0.5076)(1.203,0.5108)(1.204,0.5124)(1.205,0.5141)(1.206,0.5157)(1.208,0.5189)(1.209,0.5205)(1.21,0.5221)(1.211,0.5237)(1.213,0.5269)(1.214,0.5284)(1.215,0.5300)(1.216,0.5316)(1.218,0.5348)(1.219,0.5364)(1.22,0.5380)(1.221,0.5396)(1.223,0.5428)(1.224,0.5444)(1.225,0.5460)(1.226,0.5476)(1.228,0.5508)(1.229,0.5524)(1.23,0.5540)(1.231,0.5555)(1.233,0.5587)(1.234,0.5603)(1.235,0.5619)(1.236,0.5635)(1.238,0.5666)(1.239,0.5682)(1.24,0.5698)(1.242,0.5730)(1.243,0.5745)(1.244,0.5761)(1.245,0.5777)(1.247,0.5808)(1.248,0.5824)(1.249,0.5840)(1.25,0.5856)(1.252,0.5887)(1.253,0.5903)(1.254,0.5918)(1.256,0.5949)(1.257,0.5965)(1.258,0.5981)(1.259,0.5996)(1.261,0.6027)(1.262,0.6043)(1.263,0.6058)(1.265,0.6089)(1.266,0.6105)(1.267,0.6120)(1.269,0.6151)(1.27,0.6166)(1.271,0.6182)(1.272,0.6197)(1.274,0.6228)(1.275,0.6243)(1.276,0.6258)(1.278,0.6289)(1.279,0.6304)(1.28,0.6319)(1.282,0.6350)(1.283,0.6365)(1.284,0.6380)(1.286,0.6410)(1.287,0.6425)(1.288,0.6440)(1.29,0.6470)(1.291,0.6485)(1.292,0.6500)(1.294,0.6530)(1.295,0.6545)(1.296,0.6560)(1.298,0.6590)(1.299,0.6604)(1.3,0.6619)(1.302,0.6649)(1.303,0.6663)(1.304,0.6678)(1.306,0.6707)(1.307,0.6722)(1.309,0.6751)(1.31,0.6766)(1.311,0.6780)(1.313,0.6809)(1.314,0.6823)(1.316,0.6852)(1.317,0.6867)(1.318,0.6881)(1.32,0.6910)(1.321,0.6924)(1.323,0.6952)(1.324,0.6966)(1.325,0.6980)(1.327,0.7009)(1.328,0.7023)(1.33,0.7051)(1.331,0.7065)(1.333,0.7092)(1.334,0.7106)(1.335,0.7120)(1.337,0.7148)(1.338,0.7161)(1.34,0.7189)(1.341,0.7202)(1.343,0.7230)(1.344,0.7243)(1.346,0.7270)(1.347,0.7284)(1.349,0.7311)(1.35,0.7324)(1.352,0.7351)(1.353,0.7364)(1.355,0.7390)(1.356,0.7403)(1.358,0.7430)(1.359,0.7443)(1.361,0.7469)(1.362,0.7482)(1.364,0.7508)(1.365,0.7520)(1.367,0.7546)(1.369,0.7572)(1.37,0.7584)(1.372,0.7610)(1.373,0.7622)(1.375,0.7647)(1.377,0.7672)(1.378,0.7685)(1.38,0.7709)(1.381,0.7722)(1.383,0.7746)(1.385,0.7770)(1.386,0.7783)(1.388,0.7807)(1.39,0.7831)(1.391,0.7843)(1.393,0.7866)(1.395,0.7890)(1.396,0.7902)(1.398,0.7925)(1.4,0.7949)(1.402,0.7972)(1.403,0.7983)(1.405,0.8006)(1.407,0.8029)(1.409,0.8051)(1.41,0.8063)(1.412,0.8085)(1.414,0.8107)(1.416,0.8129)(1.418,0.8151)(1.419,0.8162)(1.421,0.8184)(1.423,0.8205)(1.425,0.8227)(1.427,0.8248)(1.429,0.8269)(1.431,0.8290)(1.433,0.8311)(1.435,0.8331)(1.437,0.8352)(1.439,0.8372)(1.441,0.8392)(1.443,0.8412)(1.445,0.8432)(1.447,0.8452)(1.449,0.8471)(1.451,0.8491)(1.453,0.8510)(1.455,0.8529)(1.457,0.8548)(1.459,0.8566)(1.461,0.8585)(1.463,0.8603)(1.466,0.8631)(1.468,0.8649)(1.47,0.8667)(1.472,0.8684)(1.475,0.8711)(1.477,0.8728)(1.479,0.8745)(1.482,0.8770)(1.484,0.8787)(1.486,0.8804)(1.489,0.8829)(1.491,0.8845)(1.494,0.8869)(1.496,0.8885)(1.499,0.8908)(1.501,0.8924)(1.504,0.8947)(1.507,0.8969)(1.509,0.8984)(1.512,0.9006)(1.515,0.9028)(1.518,0.9049)(1.52,0.9064)(1.523,0.9084)(1.526,0.9105)(1.529,0.9125)(1.532,0.9145)(1.535,0.9165)(1.538,0.9184)(1.541,0.9203)(1.545,0.9227)(1.548,0.9246)(1.551,0.9264)(1.555,0.9287)(1.558,0.9304)(1.562,0.9327)(1.565,0.9343)(1.569,0.9365)(1.573,0.9386)(1.577,0.9406)(1.581,0.9427)(1.585,0.9446)(1.589,0.9465)(1.593,0.9484)(1.597,0.9502)(1.602,0.9524)(1.606,0.9541)(1.611,0.9562)(1.616,0.9582)(1.621,0.9601)(1.627,0.9624)(1.632,0.9642)(1.638,0.9662)(1.644,0.9682)(1.65,0.9701)(1.656,0.9719)(1.663,0.9740)(1.67,0.9759)(1.678,0.9779)(1.686,0.9799)(1.694,0.9817)(1.704,0.9838)(1.713,0.9855)(1.724,0.9875)(1.736,0.9894)(1.749,0.9913)(1.764,0.9932)(1.782,0.9951)(1.802,0.9968)(1.828,0.9984)(1.863,0.9997)(1.92,1.0000)
      };
      
      \addplot [darkgreen, line width=0.6pt, opacity=0.4] coordinates {
        (0.48,0.0)(0.782,0.0)
        (0.782,0.1)(0.936,0.1)
        (0.936,0.2)(1.028,0.2)
        (1.028,0.3)(1.100,0.3)
        (1.100,0.4)(1.165,0.4)
        (1.165,0.5)(1.228,0.5)
        (1.228,0.6)(1.299,0.6)
        (1.299,0.7)(1.364,0.7)
        (1.364,0.8)(1.452,0.8)
        (1.452,0.9)(1.597,0.9)
        (1.597,1.0)(1.920,1.0)
      };

      \addplot [densely dashed, darkred, smooth, line width=0.6pt] coordinates {
        (0.903,0.00360)(1.003,0.00619)(1.035,0.00733)(1.06,0.00850)(1.079,0.00958)(1.095,0.01063)(1.109,0.01169)(1.122,0.01279)(1.133,0.01382)(1.143,0.01485)(1.152,0.01585)(1.161,0.01694)(1.169,0.01796)(1.176,0.01892)(1.183,0.01994)(1.19,0.02101)(1.196,0.02198)(1.203,0.02316)(1.208,0.02405)(1.214,0.02516)(1.219,0.02613)(1.224,0.02713)(1.229,0.02817)(1.234,0.02924)(1.238,0.03014)(1.243,0.03129)(1.247,0.03224)(1.251,0.03322)(1.255,0.03423)(1.259,0.03527)(1.263,0.03633)(1.266,0.03716)(1.27,0.03828)(1.273,0.03914)(1.277,0.04032)(1.28,0.04123)(1.283,0.04215)(1.287,0.04342)(1.29,0.04438)(1.293,0.04537)(1.296,0.04638)(1.299,0.04740)(1.302,0.04845)(1.304,0.04916)(1.307,0.05025)(1.31,0.05135)(1.313,0.05248)(1.315,0.05324)(1.318,0.05441)(1.32,0.05520)(1.323,0.05640)(1.325,0.05721)(1.328,0.05845)(1.33,0.05929)(1.333,0.06057)(1.335,0.06143)(1.337,0.06231)(1.339,0.06320)(1.342,0.06455)(1.344,0.06546)(1.346,0.06639)(1.348,0.06733)(1.35,0.06828)(1.352,0.06924)(1.354,0.07022)(1.356,0.07120)(1.358,0.07220)(1.36,0.07321)(1.362,0.07423)(1.364,0.07526)(1.366,0.07631)(1.368,0.07736)(1.37,0.07843)(1.372,0.07951)(1.374,0.08060)(1.376,0.08170)(1.377,0.08226)(1.379,0.08338)(1.381,0.08452)(1.383,0.08566)(1.384,0.08624)(1.386,0.08741)(1.388,0.08859)(1.389,0.08919)(1.391,0.09039)(1.393,0.09160)(1.394,0.09222)(1.396,0.09345)(1.398,0.09470)(1.399,0.09533)(1.401,0.09660)(1.402,0.09723)(1.404,0.09852)(1.405,0.09917)(1.407,0.10048)(1.408,0.10114)(1.41,0.10248)(1.411,0.10315)(1.413,0.10450)(1.414,0.10518)(1.416,0.10656)(1.417,0.10725)(1.419,0.10864)(1.42,0.10934)(1.422,0.11076)(1.423,0.11147)(1.424,0.11219)(1.426,0.11364)(1.427,0.11436)(1.428,0.11510)(1.43,0.11657)(1.431,0.11731)(1.433,0.11881)(1.434,0.11956)(1.435,0.12032)(1.436,0.12108)(1.438,0.12262)(1.439,0.12339)(1.44,0.12417)(1.442,0.12574)(1.443,0.12652)(1.444,0.12732)(1.445,0.12811)(1.447,0.12972)(1.448,0.13052)(1.449,0.13133)(1.45,0.13215)(1.452,0.13379)(1.453,0.13462)(1.454,0.13545)(1.455,0.13629)(1.456,0.13713)(1.458,0.13881)(1.459,0.13966)(1.46,0.14052)(1.461,0.14138)(1.462,0.14224)(1.463,0.14311)(1.465,0.14485)(1.466,0.14573)(1.467,0.14661)(1.468,0.14750)(1.469,0.14839)(1.47,0.14928)(1.471,0.15019)(1.472,0.15109)(1.473,0.15200)(1.475,0.15382)(1.476,0.15474)(1.477,0.15567)(1.478,0.15659)(1.479,0.15752)(1.48,0.15846)(1.481,0.15940)(1.482,0.16034)(1.483,0.16129)(1.484,0.16225)(1.485,0.16320)(1.486,0.16417)(1.487,0.16513)(1.488,0.16610)(1.489,0.16708)(1.49,0.16806)(1.491,0.16904)(1.492,0.17003)(1.493,0.17102)(1.494,0.17201)(1.495,0.17302)(1.496,0.17402)(1.497,0.17503)(1.498,0.17604)(1.499,0.17706)(1.5,0.17808)(1.501,0.17911)(1.502,0.18014)(1.503,0.18117)(1.504,0.18221)(1.505,0.18325)(1.506,0.18429)(1.507,0.18534)(1.508,0.18640)(1.509,0.18746)(1.51,0.18852)(1.511,0.18959)(1.512,0.19066)(1.513,0.19173)(1.514,0.19281)(1.514,0.19281)(1.515,0.19390)(1.516,0.19499)(1.517,0.19608)(1.518,0.19718)(1.519,0.19828)(1.52,0.19939)(1.521,0.20050)(1.522,0.20161)(1.523,0.20273)(1.524,0.20385)(1.524,0.20385)(1.525,0.20498)(1.526,0.20611)(1.527,0.20725)(1.528,0.20839)(1.529,0.20953)(1.53,0.21068)(1.531,0.21183)(1.531,0.21183)(1.532,0.21299)(1.533,0.21416)(1.534,0.21532)(1.535,0.21649)(1.536,0.21767)(1.537,0.21885)(1.537,0.21885)(1.538,0.22003)(1.539,0.22122)(1.54,0.22241)(1.541,0.22361)(1.542,0.22481)(1.542,0.22481)(1.543,0.22602)(1.544,0.22723)(1.545,0.22844)(1.546,0.22966)(1.546,0.22966)(1.547,0.23088)(1.548,0.23211)(1.549,0.23334)(1.55,0.23458)(1.551,0.23582)(1.551,0.23582)(1.552,0.23707)(1.553,0.23832)(1.554,0.23957)(1.555,0.24082)(1.555,0.24082)(1.556,0.24209)(1.557,0.24336)(1.558,0.24463)(1.558,0.24463)(1.559,0.24590)(1.56,0.24718)(1.561,0.24847)(1.562,0.24976)(1.562,0.24976)(1.563,0.25105)(1.564,0.25235)(1.565,0.25365)(1.565,0.25365)(1.566,0.25496)(1.567,0.25627)(1.568,0.25758)(1.568,0.25758)(1.569,0.25890)(1.57,0.26022)(1.571,0.26155)(1.571,0.26155)(1.572,0.26288)(1.573,0.26422)(1.574,0.26556)(1.574,0.26556)(1.575,0.26691)(1.576,0.26826)(1.577,0.26961)(1.577,0.26961)(1.578,0.27097)(1.579,0.27233)(1.58,0.27370)(1.58,0.27370)(1.581,0.27507)(1.582,0.27644)(1.582,0.27644)(1.583,0.27782)(1.584,0.27921)(1.585,0.28059)(1.585,0.28059)(1.586,0.28199)(1.587,0.28338)(1.588,0.28478)(1.588,0.28478)(1.589,0.28619)(1.59,0.28760)(1.59,0.28760)(1.591,0.28901)(1.592,0.29043)(1.592,0.29043)(1.593,0.29185)(1.594,0.29328)(1.595,0.29471)(1.595,0.29471)(1.596,0.29614)(1.597,0.29758)(1.597,0.29758)(1.598,0.29902)(1.599,0.30047)(1.599,0.30047)(1.6,0.30192)(1.601,0.30337)(1.601,0.30337)(1.602,0.30483)(1.603,0.30630)(1.604,0.30776)(1.604,0.30776)(1.605,0.30923)(1.606,0.31071)(1.606,0.31071)(1.607,0.31219)(1.608,0.31367)(1.608,0.31367)(1.609,0.31516)(1.61,0.31665)(1.61,0.31665)(1.611,0.31814)(1.612,0.31964)(1.612,0.31964)(1.613,0.32115)(1.614,0.32265)(1.614,0.32265)(1.615,0.32416)(1.616,0.32568)(1.616,0.32568)(1.617,0.32720)(1.617,0.32720)(1.618,0.32872)(1.619,0.33025)(1.619,0.33025)(1.62,0.33178)(1.621,0.33332)(1.621,0.33332)(1.622,0.33486)(1.623,0.33640)(1.623,0.33640)(1.624,0.33795)(1.625,0.33950)(1.625,0.33950)(1.626,0.34106)(1.627,0.34262)(1.627,0.34262)(1.628,0.34418)(1.628,0.34418)(1.629,0.34575)(1.63,0.34732)(1.63,0.34732)(1.631,0.34889)(1.632,0.35047)(1.632,0.35047)(1.633,0.35205)(1.634,0.35364)(1.634,0.35364)(1.635,0.35523)(1.635,0.35523)(1.636,0.35682)(1.637,0.35842)(1.637,0.35842)(1.638,0.36002)(1.639,0.36162)(1.639,0.36162)(1.64,0.36323)(1.64,0.36323)(1.641,0.36484)(1.642,0.36646)(1.642,0.36646)(1.643,0.36807)(1.643,0.36807)(1.644,0.36970)(1.645,0.37132)(1.645,0.37132)(1.646,0.37295)(1.647,0.37459)(1.647,0.37459)(1.648,0.37622)(1.648,0.37622)(1.649,0.37786)(1.65,0.37951)(1.65,0.37951)(1.651,0.38115)(1.651,0.38115)(1.652,0.38281)(1.653,0.38446)(1.653,0.38446)(1.654,0.38612)(1.654,0.38612)(1.655,0.38778)(1.656,0.38944)(1.656,0.38944)(1.657,0.39111)(1.657,0.39111)(1.658,0.39278)(1.659,0.39446)(1.659,0.39446)(1.66,0.39614)(1.66,0.39614)(1.661,0.39782)(1.662,0.39950)(1.662,0.39950)(1.663,0.40119)(1.663,0.40119)(1.664,0.40288)(1.664,0.40288)(1.665,0.40458)(1.666,0.40627)(1.666,0.40627)(1.667,0.40797)(1.667,0.40797)(1.668,0.40968)(1.669,0.41139)(1.669,0.41139)(1.67,0.41310)(1.67,0.41310)(1.671,0.41481)(1.671,0.41481)(1.672,0.41653)(1.673,0.41825)(1.673,0.41825)(1.674,0.41997)(1.674,0.41997)(1.675,0.42169)(1.676,0.42342)(1.676,0.42342)(1.677,0.42515)(1.677,0.42515)(1.678,0.42689)(1.678,0.42689)(1.679,0.42863)(1.68,0.43037)(1.68,0.43037)(1.681,0.43211)(1.681,0.43211)(1.682,0.43386)(1.682,0.43386)(1.683,0.43561)(1.684,0.43736)(1.684,0.43736)(1.685,0.43911)(1.685,0.43911)(1.686,0.44087)(1.686,0.44087)(1.687,0.44263)(1.688,0.44439)(1.688,0.44439)(1.689,0.44616)(1.689,0.44616)(1.69,0.44793)(1.69,0.44793)(1.691,0.44970)(1.691,0.44970)(1.692,0.45147)(1.693,0.45325)(1.693,0.45325)(1.694,0.45503)(1.694,0.45503)(1.695,0.45681)(1.695,0.45681)(1.696,0.45859)(1.697,0.46038)(1.697,0.46038)(1.698,0.46217)(1.698,0.46217)(1.699,0.46396)(1.699,0.46396)(1.7,0.46575)(1.7,0.46575)(1.701,0.46755)(1.702,0.46935)(1.702,0.46935)(1.703,0.47115)(1.703,0.47115)(1.704,0.47295)(1.704,0.47295)(1.705,0.47475)(1.705,0.47475)(1.706,0.47656)(1.706,0.47656)(1.707,0.47837)(1.708,0.48018)(1.708,0.48018)(1.709,0.48200)(1.709,0.48200)(1.71,0.48381)(1.71,0.48381)(1.711,0.48563)(1.711,0.48563)(1.712,0.48745)(1.713,0.48927)(1.713,0.48927)(1.714,0.49110)(1.714,0.49110)(1.715,0.49292)(1.715,0.49292)(1.716,0.49475)(1.716,0.49475)(1.717,0.49658)(1.717,0.49658)(1.718,0.49842)(1.719,0.50025)(1.719,0.50025)(1.72,0.50208)(1.72,0.50208)(1.721,0.50392)(1.721,0.50392)(1.722,0.50576)(1.722,0.50576)(1.723,0.50760)(1.723,0.50760)(1.724,0.50944)(1.724,0.50944)(1.725,0.51129)(1.726,0.51314)(1.726,0.51314)(1.727,0.51498)(1.727,0.51498)(1.728,0.51683)(1.728,0.51683)(1.729,0.51868)(1.729,0.51868)(1.73,0.52054)(1.73,0.52054)(1.731,0.52239)(1.731,0.52239)(1.732,0.52425)(1.733,0.52610)(1.733,0.52610)(1.734,0.52796)(1.734,0.52796)(1.735,0.52982)(1.735,0.52982)(1.736,0.53168)(1.736,0.53168)(1.737,0.53354)(1.737,0.53354)(1.738,0.53541)(1.738,0.53541)(1.739,0.53727)(1.739,0.53727)(1.74,0.53914)(1.741,0.54100)(1.741,0.54100)(1.742,0.54287)(1.742,0.54287)(1.743,0.54474)(1.743,0.54474)(1.744,0.54661)(1.744,0.54661)(1.745,0.54848)(1.745,0.54848)(1.746,0.55036)(1.746,0.55036)(1.747,0.55223)(1.747,0.55223)(1.748,0.55410)(1.749,0.55598)(1.749,0.55598)(1.75,0.55785)(1.75,0.55785)(1.751,0.55973)(1.751,0.55973)(1.752,0.56161)(1.752,0.56161)(1.753,0.56348)(1.753,0.56348)(1.754,0.56536)(1.754,0.56536)(1.755,0.56724)(1.755,0.56724)(1.756,0.56912)(1.756,0.56912)(1.757,0.57100)(1.758,0.57289)(1.758,0.57289)(1.759,0.57477)(1.759,0.57477)(1.76,0.57665)(1.76,0.57665)(1.761,0.57853)(1.761,0.57853)(1.762,0.58041)(1.762,0.58041)(1.763,0.58230)(1.763,0.58230)(1.764,0.58418)(1.764,0.58418)(1.765,0.58606)(1.765,0.58606)(1.766,0.58795)(1.767,0.58983)(1.767,0.58983)(1.768,0.59172)(1.768,0.59172)(1.769,0.59360)(1.769,0.59360)(1.77,0.59549)(1.77,0.59549)(1.771,0.59737)(1.771,0.59737)(1.772,0.59926)(1.772,0.59926)(1.773,0.60114)(1.773,0.60114)(1.774,0.60303)(1.774,0.60303)(1.775,0.60491)(1.776,0.60679)(1.776,0.60679)(1.777,0.60868)(1.777,0.60868)(1.778,0.61056)(1.778,0.61056)(1.779,0.61245)(1.779,0.61245)(1.78,0.61433)(1.78,0.61433)(1.781,0.61621)(1.781,0.61621)(1.782,0.61809)(1.782,0.61809)(1.783,0.61998)(1.783,0.61998)(1.784,0.62186)(1.785,0.62374)(1.785,0.62374)(1.786,0.62562)(1.786,0.62562)(1.787,0.62750)(1.787,0.62750)(1.788,0.62938)(1.788,0.62938)(1.789,0.63126)(1.789,0.63126)(1.79,0.63314)(1.79,0.63314)(1.791,0.63501)(1.791,0.63501)(1.792,0.63689)(1.792,0.63689)(1.793,0.63876)(1.794,0.64064)(1.794,0.64064)(1.795,0.64251)(1.795,0.64251)(1.796,0.64439)(1.796,0.64439)(1.797,0.64626)(1.797,0.64626)(1.798,0.64813)(1.798,0.64813)(1.799,0.65000)(1.799,0.65000)(1.8,0.65187)(1.8,0.65187)(1.801,0.65373)(1.802,0.65560)(1.802,0.65560)(1.803,0.65747)(1.803,0.65747)(1.804,0.65933)(1.804,0.65933)(1.805,0.66119)(1.805,0.66119)(1.806,0.66305)(1.806,0.66305)(1.807,0.66491)(1.807,0.66491)(1.808,0.66677)(1.808,0.66677)(1.809,0.66863)(1.81,0.67048)(1.81,0.67048)(1.811,0.67233)(1.811,0.67233)(1.812,0.67419)(1.812,0.67419)(1.813,0.67604)(1.813,0.67604)(1.814,0.67788)(1.814,0.67788)(1.815,0.67973)(1.815,0.67973)(1.816,0.68157)(1.817,0.68342)(1.817,0.68342)(1.818,0.68526)(1.818,0.68526)(1.819,0.68710)(1.819,0.68710)(1.82,0.68893)(1.82,0.68893)(1.821,0.69077)(1.821,0.69077)(1.822,0.69260)(1.823,0.69443)(1.823,0.69443)(1.824,0.69626)(1.824,0.69626)(1.825,0.69809)(1.825,0.69809)(1.826,0.69991)(1.826,0.69991)(1.827,0.70173)(1.827,0.70173)(1.828,0.70355)(1.829,0.70537)(1.829,0.70537)(1.83,0.70718)(1.83,0.70718)(1.831,0.70900)(1.831,0.70900)(1.832,0.71080)(1.832,0.71080)(1.833,0.71261)(1.833,0.71261)(1.834,0.71441)(1.835,0.71622)(1.835,0.71622)(1.836,0.71802)(1.836,0.71802)(1.837,0.71981)(1.837,0.71981)(1.838,0.72161)(1.838,0.72161)(1.839,0.72340)(1.84,0.72518)(1.84,0.72518)(1.841,0.72697)(1.841,0.72697)(1.842,0.72875)(1.842,0.72875)(1.843,0.73053)(1.843,0.73053)(1.844,0.73230)(1.845,0.73408)(1.845,0.73408)(1.846,0.73585)(1.846,0.73585)(1.847,0.73761)(1.847,0.73761)(1.848,0.73938)(1.849,0.74114)(1.849,0.74114)(1.85,0.74289)(1.85,0.74289)(1.851,0.74465)(1.851,0.74465)(1.852,0.74639)(1.853,0.74814)(1.853,0.74814)(1.854,0.74989)(1.854,0.74989)(1.855,0.75162)(1.855,0.75162)(1.856,0.75336)(1.857,0.75509)(1.857,0.75509)(1.858,0.75682)(1.858,0.75682)(1.859,0.75855)(1.859,0.75855)(1.86,0.76027)(1.861,0.76198)(1.861,0.76198)(1.862,0.76370)(1.862,0.76370)(1.863,0.76541)(1.863,0.76541)(1.864,0.76711)(1.865,0.76881)(1.865,0.76881)(1.866,0.77051)(1.866,0.77051)(1.867,0.77220)(1.868,0.77389)(1.868,0.77389)(1.869,0.77558)(1.869,0.77558)(1.87,0.77726)(1.871,0.77894)(1.871,0.77894)(1.872,0.78061)(1.872,0.78061)(1.873,0.78228)(1.873,0.78228)(1.874,0.78394)(1.875,0.78560)(1.875,0.78560)(1.876,0.78726)(1.876,0.78726)(1.877,0.78891)(1.878,0.79055)(1.878,0.79055)(1.879,0.79219)(1.88,0.79383)(1.88,0.79383)(1.881,0.79546)(1.881,0.79546)(1.882,0.79709)(1.883,0.79871)(1.883,0.79871)(1.884,0.80033)(1.884,0.80033)(1.885,0.80195)(1.886,0.80355)(1.886,0.80355)(1.887,0.80516)(1.888,0.80676)(1.888,0.80676)(1.889,0.80835)(1.889,0.80835)(1.89,0.80994)(1.891,0.81152)(1.891,0.81152)(1.892,0.81310)(1.893,0.81468)(1.893,0.81468)(1.894,0.81625)(1.894,0.81625)(1.895,0.81781)(1.896,0.81937)(1.896,0.81937)(1.897,0.82092)(1.898,0.82247)(1.898,0.82247)(1.899,0.82401)(1.9,0.82554)(1.9,0.82554)(1.901,0.82708)(1.902,0.82860)(1.902,0.82860)(1.903,0.83012)(1.903,0.83012)(1.904,0.83163)(1.905,0.83314)(1.905,0.83314)(1.906,0.83465)(1.907,0.83614)(1.907,0.83614)(1.908,0.83763)(1.909,0.83912)(1.909,0.83912)(1.91,0.84060)(1.911,0.84207)(1.911,0.84207)(1.912,0.84354)(1.913,0.84500)(1.914,0.84646)(1.914,0.84646)(1.915,0.84791)(1.916,0.84936)(1.916,0.84936)(1.917,0.85080)(1.918,0.85223)(1.918,0.85223)(1.919,0.85366)(1.92,0.85508)(1.92,0.85508)(1.921,0.85649)(1.922,0.85790)(1.923,0.85930)(1.923,0.85930)(1.924,0.86070)(1.925,0.86209)(1.925,0.86209)(1.926,0.86347)(1.927,0.86485)(1.928,0.86622)(1.928,0.86622)(1.929,0.86758)(1.93,0.86894)(1.931,0.87030)(1.931,0.87030)(1.932,0.87164)(1.933,0.87298)(1.933,0.87298)(1.934,0.87431)(1.935,0.87563)(1.936,0.87695)(1.936,0.87695)(1.937,0.87826)(1.938,0.87957)(1.939,0.88087)(1.94,0.88216)(1.94,0.88216)(1.941,0.88344)(1.942,0.88472)(1.943,0.88599)(1.943,0.88599)(1.944,0.88726)(1.945,0.88851)(1.946,0.88977)(1.947,0.89101)(1.947,0.89101)(1.948,0.89225)(1.949,0.89348)(1.95,0.89470)(1.951,0.89592)(1.951,0.89592)(1.952,0.89712)(1.953,0.89832)(1.954,0.89952)(1.955,0.90071)(1.956,0.90189)(1.956,0.90189)(1.957,0.90306)(1.958,0.90423)(1.959,0.90538)(1.96,0.90654)(1.961,0.90768)(1.962,0.90882)(1.962,0.90882)(1.963,0.90995)(1.964,0.91107)(1.965,0.91218)(1.966,0.91329)(1.967,0.91439)(1.968,0.91548)(1.969,0.91657)(1.97,0.91765)(1.971,0.91873)(1.972,0.91979)(1.972,0.91979)(1.973,0.92085)(1.974,0.92189)(1.975,0.92293)(1.976,0.92396)(1.977,0.92499)(1.978,0.92601)(1.979,0.92702)(1.98,0.92802)(1.981,0.92901)(1.982,0.93000)(1.983,0.93098)(1.984,0.93195)(1.985,0.93292)(1.986,0.93388)(1.987,0.93482)(1.988,0.93577)(1.989,0.93670)(1.991,0.93855)(1.992,0.93947)(1.993,0.94037)(1.994,0.94127)(1.995,0.94217)(1.996,0.94305)(1.997,0.94392)(1.998,0.94479)(1.999,0.94565)(2.001,0.94734)(2.002,0.94818)(2.003,0.94901)(2.004,0.94983)(2.005,0.95065)(2.007,0.95225)(2.008,0.95304)(2.009,0.95383)(2.01,0.95460)(2.012,0.95613)(2.013,0.95688)(2.014,0.95763)(2.016,0.95909)(2.017,0.95982)(2.019,0.96124)(2.02,0.96195)(2.021,0.96264)(2.023,0.96400)(2.024,0.96467)(2.026,0.96599)(2.027,0.96664)(2.029,0.96791)(2.03,0.96854)(2.032,0.96977)(2.034,0.97097)(2.035,0.97156)(2.037,0.97271)(2.039,0.97384)(2.041,0.97495)(2.042,0.97549)(2.044,0.97654)(2.046,0.97756)(2.048,0.97856)(2.05,0.97953)(2.052,0.98047)(2.054,0.98139)(2.057,0.98272)(2.059,0.98357)(2.061,0.98439)(2.064,0.98558)(2.066,0.98634)(2.069,0.98742)(2.072,0.98846)(2.075,0.98944)(2.078,0.99037)(2.081,0.99125)(2.085,0.99233)(2.089,0.99333)(2.093,0.99425)(2.097,0.99507)(2.102,0.99598)(2.108,0.99692)(2.115,0.99781)(2.124,0.99868)(2.137,0.99944)(2.155,1.00000)(2.2,1.00000)
      };
      
      \addplot [darkred, line width=0.6pt, opacity=0.4] coordinates {
        (0.903,0.0)(1.307,0.0)
        (1.307,0.1)(1.471,0.1)
        (1.471,0.2)(1.562,0.2)
        (1.562,0.3)(1.632,0.3)
        (1.632,0.4)(1.691,0.4)
        (1.691,0.5)(1.746,0.5)
        (1.746,0.6)(1.799,0.6)
        (1.799,0.7)(1.854,0.7)
        (1.854,0.8)(1.917,0.8)
        (1.917,0.9)(2.005,0.9)
        (2.005,1.0)(2.2,1.0)
      };
      
      \legend{$\obj(A)$, , $\obj(B)$, }
      
    \end{axis}
  \end{scope}
\end{tikzpicture}
}

\begin{document}

\title{Efficient Computation of Probabilistic Dominance\\ in Robust Multi-Objective Optimization}

\title{Efficient Computation of Probabilistic Dominance\\ in Robust Multi-Objective Optimization
\thanks{ The authors are with the Department of Computer Science, Friedrich-Alexander-Universit\"at Erlangen-N\"urnberg (FAU), Erlangen 91058, Germany.
(e-mail: \{faramarz.khosravi, alexander.rass, juergen.teich\}@fau.de).
}%
}
\author{Faramarz Khosravi, Alexander Ra\ss, and J\"urgen Teich,}

\maketitle 

\begin{abstract}
Real-world problems typically require the simultaneous optimization of several, often conflicting objectives.
Many of these \emph{multi-objective optimization problems} are characterized by wide ranges of uncertainties in their decision variables or objective functions, which further increases the complexity of optimization.
To cope with such uncertainties, \emph{robust optimization} is widely studied aiming to distinguish candidate solutions with uncertain objectives specified by confidence intervals, probability distributions or sampled data.
However, existing techniques mostly either fail to consider the actual distributions or assume uncertainty as instances of uniform or Gaussian distributions.
This paper introduces an empirical approach that enables an efficient comparison of candidate solutions with uncertain objectives that can follow arbitrary distributions.
Given two candidate solutions under comparison, this operator calculates the probability that one solution dominates the other in terms of each uncertain objective.
It can substitute for the standard comparison operator of existing optimization techniques such as evolutionary algorithms to enable discovering robust solutions to problems with multiple uncertain objectives.
This paper also proposes to incorporate various uncertainties in well-known multi-objective problems to provide a benchmark for evaluating uncertainty-aware optimization techniques. 
The proposed comparison operator and benchmark suite are integrated into an existing optimization tool that features a selection of multi-objective optimization problems and algorithms.
Experiments show that in comparison with existing techniques, the proposed approach achieves higher optimization quality at lower overheads.

\end{abstract}

\acrodef{CDF}{cumulative distribution function}
\acrodef{DCI}{diversity comparison indicator}
\acrodef{PDF}{probability density function}

\section*{Notation}

\begin{description}
 \item[$A$] a candidate solution
 \item[$B$] a candidate solution
 \item[$\BetaNoPar$] beta distribution
 \item[$C$] comparison operator
 \item[$c$] positive constant
 \item[$E$] expected value
 \item[$e$] approximation error
 \item[$\obj$] objective function
 \item[$\GaussNoPar$] Gaussian distribution
 \item[$N$] number of samples or quantile cuts
 \item[$n$] number of decision variables
 \item[$m$] number of objective functions
 \item[$S$] sequence of samples
 \item[$s$] sample from an uncertain objective's distribution
 \item[$\UniformNoPar$] uniform distribution
 \item[$\bm{u}$] uncertainty added to an optimization problem
 \item[$Var$] variance
 \item[$X$] random variable
 \item[$x$] decision variable
 \item[$\gamma$] comparison threshold
 \item[$\delta$] tolerance (bound on an error)
 \item[$\sigma$] standard deviation
 \item[$\omega$] interval width in a histogram
\end{description}

\section{Introduction}\label{sec:introduction}
Real-world problems typically demand solutions that are optimized with respect to multiple criteria called objectives.
In these so-called \emph{multi-objective optimization problems}, the objectives often conflict with each other such that no single solution can be found to be optimal in all objectives.
Instead, one usually searches for a set of non-dominated solutions known as \emph{Pareto front} or \emph{Pareto set} that provide decent trade-offs among objectives.
A solution is said to \emph{dominate} another if it is as good as the other in all objectives and is better with respect to at least one objective.
While exact optimization methods such as \emph{integer linear programming} may not be applicable to complex optimization problems, 
population-based \emph{meta-heuristics} such as \emph{evolutionary algorithms} enable a fast approximation to the Pareto front of problems with several objectives and large search spaces~\cite{JBT09}.

However, multi-objective optimization problems are often characterized by wide ranges of uncertainties including noise, approximation errors or time-dependent variation in their objective functions and perturbations in their decision variables~\cite{JB05}. 
Any optimization algorithm that neglects the effects of uncertainty might prefer actually inferior solutions while traversing the search space.
As a remedy, \emph{robust optimization} techniques have been proposed to enable an accurate comparison of objectives in the presence of uncertainty.
Existing techniques typically model uncertain objectives using instances of uniform~\cite{Teich01} and Gaussian~\cite{Hughes01,FE05} distributions, intervals specified by best and worst cases~\cite{Limbourg05, EM13}, or sampled data~\cite{MAG12,KMGT15}.
While the first two groups fail to deal with various, and possibly non-standard uncertainty distributions, the third group of techniques enable the comparison of candidate solutions with arbitrarily distributed uncertain objectives.
However, these techniques rely on estimated statistics such as mean value and variance and do not take the actual uncertainty distributions into account. 
This problem is addressed in a previous work of the authors~\cite{KBT18}, which enables calculating the probability that an uncertain objective of one solution is greater or smaller than that of another solution, for any arbitrary distribution given as a closed-form function or sampled data.
This probability is calculated through partitioning the probability distribution of objectives into small intervals of the same size and applying rectangle integration, \ie Riemann sum, while assuming a uniform distribution within each interval. 
Therefore, the optimization algorithm can differentiate instances of each uncertain objectives of two candidate solutions under comparison, and determine whether one solution dominates the other or not. 

This paper extends the comparison operator in~\cite{KBT18} by introducing a new method for the calculation of the probability that an instance of uncertain objective is greater than another instance of the same objective.
This method is based on obtaining the \ac{CDF} of each uncertain objective, and partitioning the probability space into intervals of the same size.
It uses an iterative approach similar to~\cite{KBT18}, except that its accuracy is not impaired by the assumption of uniform distribution within intervals.

Moreover, this paper extends the well-known DTLZ multi-objective benchmark suite~\cite{DTLZ02} to consider the effects of various uncertainties. 
It integrates the proposed comparison operator and the extended benchmark into the multi-objective optimization framework Opt4J~\cite{opt4jpaper} that incorporates several optimization algorithms such as evolutionary algorithms~\cite{DAPM00} and particle swarm optimization~\cite{SC06}.
Experiments show that compared to the existing techniques, the proposed approach enables comparing uncertain objectives more efficiently and achieves higher optimization quality.

The rest of this paper is structured as follows: 
Section~\ref{sec:relatedwork} reviews the state-of-the-art techniques for robust multi-objective optimization. 
Sections~\ref{sec:proposed} and~\ref{sec:benchmark} respectively introduce the proposed comparison operator and uncertain multi-objective optimization benchmark. 
Section~\ref{sec:experiments} presents the experimental setup and evaluation results, and in the end, Section~\ref{sec:conclusion} concludes this work.

\section{Related Work}\label{sec:relatedwork}
The objective functions and decision variables of multi-objective optimization problems are often subject to various uncertainties.
In the context of probabilistic risk assessment, these uncertainties are categorized with respect to their origin as \emph{aleatory} and \emph{epistemic} uncertainties.
Aleatory uncertainty refers to ``the inherent variation associated with the physical system or the environment under consideration'', 
whereas epistemic uncertainty describes ``any lack of knowledge or information in any phase or activity of the modeling process''~\cite{OHJWF04}. 

The probability distribution of a decision variable or an objective function with aleatory uncertainty can be estimated through sampling or iterative function evaluation, respectively.
However, obtaining the exact value or probability distribution of a variable or function with epistemic uncertainty is usually impracticable or unaffordable.
In fact, only limited characteristics such as confidence intervals may be available.

To deal with epistemic uncertainty in the context of multi-objective optimization, the work in~\cite{Limbourg05} proposes to represent uncertain objective values using lower and upper bounds rather than single point estimates.
It also extends the weak and strong dominance criteria of a multi-objective evolutionary algorithm.
To balance the accuracy and execution time of the process of comparing candidate solutions, it integrates the weak dominance criterion into the process of parent selection and the strong dominance criterion into the process of updating the solution archive. 
The authors in~\cite{ERM12} model uncertainty as the lack of knowledge about the exact effects of decision variables on objective values using a triangular fuzzy representation. 
They incorporate the pessimistic, anticipated, and optimistic values of uncertain objectives into the comparison of different solutions. 
However, the proposed dominance criterion may fail to properly distinguish objective values when the intervals between pessimistic and optimistic values do not overlap, which reduces the quality or robustness of optimization.
Since the distribution of uncertainty in objective functions and decision variables are not available in the case of epistemic uncertainty, the rest of this paper focuses on dealing with aleatory uncertainty, where there is a stronger demand for its effective and efficient handling.

Another common classification of uncertainties can be found in~\cite{JB05} where the uncertainties are categorized with respect to their manifestations into four groups.
The first group includes noise in objective functions, \ie variations in the results of different evaluations of an objective function with unchanged input variables.
The second group takes perturbations in decision variables, that are the input variables of objective functions, into consideration. 
The third group describes the error of approximate objective functions which is the case when the exact evaluation of an objective function is costly or infeasible, and is therefore substituted by simulations. 
The last group models time-varying objective functions where evaluating a function with the same inputs and parameters at different points in time delivers different outputs, while the output is deterministic at any fixed point in time.
An existing uncertainty, regardless of what category it belongs to, results in objective values that should be represented by probability distributions instead of single values.
The resulting distribution may be an instance of a standard distribution such as Gaussian, or might follow any arbitrary distribution given as a \ac{PDF} or sampled data.

To enable handling uncertainty in multi-objective optimization, a group of studies~\cite{Teich01, Hughes01, FE05} propose techniques to determine \emph{probabilistic dominance} which describes the probability that one
candidate solution dominates the other.
This probability is calculated as the intersection of all probabilities that an uncertain objective value from the first solution is more favorable than the same objective of the other solution.
The techniques in~\cite{Teich01, Hughes01, FE05} are based on the simplistic assumption that different uncertain objectives are statistically independent.
Therefore, they calculate the joint probability as the product of all individual probabilities. 
Teich~\cite{Teich01} provides a mathematical approach for the calculation of probabilistic dominance given all objectives follow instances of continuous uniform distributions.
This approach can be effortlessly extended to treat uncertain objective values with any discrete distributions. 
The work in~\cite{Hughes01} assumes that each uncertain objective is affected by a Gaussian noise with known variance.
The authors in~\cite{FE05} extend this technique to enable the calculation of probabilistic dominance when instances of the same uncertain objective have the same, but unknown variance.
They propose a learning technique to reduce the number of objective function re-evaluations needed to estimate this variance.
However, the main drawback of the techniques in~\cite{Teich01, Hughes01, FE05} is that they require all uncertain objectives of a solution to be statistically independent instances
of specific distribution types.
In fact, an uncertain objective value may follow an arbitrary distribution that combines the uncertain characteristics of different decision variables. 
Also, two different objective functions sharing one or more uncertain decision variables would have statistically dependent uncertainty distributions.

The work in~\cite{EGB07} compares candidate solutions with respect to the mean values of their uncertain objectives using a strict dominance criterion.
It proposes to deal with Gaussian noise in objective values while ranking the dominated solutions in the process of parent selection of a genetic algorithm.
For each dominated solution, a strength value is calculated which is the sum of the probabilities that this solution dominates any other solution from the population.
These probabilities are calculated similar to the approach in~\cite{Hughes01}.
Each dominated solution is then ranked with respect to the difference between the sum of strength values of all solutions it dominates and that of all solutions dominating it.
The calculation of this criterion is very time-consuming and the main dominance criterion does not incorporate uncertainty in the comparison of candidate solutions. 

Another group of studies in~\cite{MAG12} and~\cite{TC11} proposes to replace each uncertain objective by one or more single-valued objectives, each representing a unique statistic such as mean or variance of the original objective.
This eliminates the need for incorporating the effects of uncertainty in the comparison operator or dominance criteria of optimization algorithms.
As an example, the work in~\cite{TC11} adopts the mean-variance model~\cite{Markowitz52} to replace the uncertain objective in a single-objective optimization with two separate objectives representing its mean and variance.
It then uses integer programming for maximizing the mean and minimizing the variance.
The main disadvantage of this approach is that it may recognize a solution with a significantly inferior mean but a slightly better variance as non-dominated, which can crowd the solution archive and slow down the optimization.
On the other hand, the technique in~\cite{MAG12} represents each uncertain objective with a single statistic which is selected based on the criticality of the objective.
For example, it uses the fifth percentiles for critical objectives which demand a high degree of robustness and mean values for the non-critical ones.
However, this technique often fails to accurately compare uncertain objectives because a single statistic cannot describe all properties of the underlying probability distributions.

The authors in~\cite{KMGT15} and~\cite{MTF14} propose to extend the operators used in existing multi-objective optimization techniques to enable coping with uncertain objectives. 
The work in~\cite{MTF14} assumes that the uncertain objectives are specified by mean values and confidence intervals.
It checks whether the confidence intervals of none of the uncertain objectives in two candidate solutions are overlapping.
In this case, it can be easily determined if one solution dominates the other or if the solutions are incomparable.
However, if the confidence intervals overlap for at least one objective, it performs an iterative reduction of confidence intervals by re-evaluating the corresponding objective functions. 
This process is continued until the intervals are no longer overlapping or no further reduction is possible.
In the latter case, the overlapping intervals are simply compared with respect to their mean values.
The work in~\cite{KMGT15} proposes to compare instances of each uncertain objective in a three-stage algorithm.
To compare two instances of an uncertain objective, this algorithm first checks if the worst-case of one is better than the best-case of the other. 
If no preference can be found, it prefers the objective value which is significantly better with respect to the mean values.
If the mean values are not sufficiently different, the algorithm checks if one objective value has a noticeably smaller deviation.
Two objective values that cannot be differentiated by any of these three comparisons are considered equal.
The comparison operators in~\cite{KMGT15} and~\cite{MTF14} enable comparing arbitrarily distributed uncertain objectives.
Moreover, the optimization algorithms employing these operators compare instances of each objective individually, which implicitly takes possible statistical dependencies among objectives into consideration. 
Nonetheless, these techniques depend on estimated statistical properties and do not reflect the probability that one candidate solution (or objective value) dominates the other.

A different approach is proposed in~\cite{HYY18} which proposes to first solve the optimization problem without the consideration of uncertainty using the multi-objective evolutionary algorithm presented in~\cite{DAPM00}.
It performs the decomposition proposed in~\cite{ZL07} to partition the objective space and represent each sub-space by a weighted sum of objectives.
Then, it maps each solution to a weighted sum such that the distance between each solution and its corresponding weighted sum is minimized.
To deal with uncertainty, it iteratively evaluates the optimal solution of each weighted sum and derives the mean and worst-case objective values. 
In the end, it removes the non-optimal solutions that are dominated by this worst-case objective value and looks for robust solutions in the neighborhood of the optimal solution of each weighted sum.
Although this technique helps identifying the robust regions~\cite{DG06} in the search space, it lacks efficiency and treats uncertainty as worst-case objective values.

To overcome the aforementioned limitations of existing uncertainty-aware optimization techniques, a histogram-based comparison operator has been proposed in~\cite{KBT18}. 
It first partitions the probability distribution of uncertain objectives into intervals of identical width.
Then, considering a uniform distribution within each interval, it calculates the probability that an instance of an uncertain objective is greater or smaller than another instance of the same objective, and enables to differentiate the two solutions with respect to this uncertain objective.
Similar to the techniques in~\cite{KMGT15} and~\cite{MTF14}, it is capable of handling problems with statistically dependent uncertain objectives because it compares instances of each objective separately.
However, at a reasonable performance overhead, it allows for considering the entire probability distribution rather than a certain number of statistics. 

In this paper, we propose an extension to the approach in~\cite{KBT18}, aiming at improving its comparison accuracy and execution time.
To represent the probability distribution of an uncertain objective, it uses a \ac{CDF} rather than a histogram.
Given a set of samples obtained from iterative evaluation of an uncertain objective, it constructs the \ac{CDF} of the corresponding uncertain objective value by sorting the samples.
For a given value of the distribution, its cumulative probability equals the proportion of samples smaller than this value to the total number of samples. 
We then introduce a fast algorithm to compare \acp{CDF} of two instances of an uncertain objective in order to calculate the probability that one is greater than the other. 
Moreover, we propose an approximate representation of \acp{CDF} which helps to significantly reduce the time complexity of this algorithm.

\section{Proposed Robust Multi-Objective Optimization}\label{sec:proposed}
A multi-objective optimization problem includes a vector of $n$ decision variables $x = (x_1, x_2,$ $\dots, x_n)$ and a vector of $m$ objective functions $f(x) = \big(f_1(x), f_2(x), \dots, f_m(x)\big)$. 
The former describes a feasible solution in the constrained search space of the problem, and the latter evaluates this solution with respect to different objectives, \ie quality metrics, that are to be maximized or minimized.
Finding a solution that is optimal in all objectives is often impossible due to the conflict between different objectives.
Therefore, multi-objective optimization algorithms typically search for a set of non-dominated solutions that offer decent tradeoffs for the conflicting objectives.
A solution $A$ dominates another solution $B$, \ie $A \succ B$, if and only if $A$ is as good as $B$ for all objectives and there is at least one objective for which $A$ is better than $B$. 
In a maximization problem, a multi-objective dominance criterion can be defined as follows:
\begin{align}
  A \succ B \iff \mathop{\forall}_{i=1}^{m} \obj_i(A) \geq \obj_i(B) \wedge \mathop{\exists}_{j=1}^{m} \obj_j(A) > \obj_j(B) \label{eq:uncertainty_optimization_dominance}
\end{align}
where $\wedge$ denotes the logical AND operation.
In the presence of uncertainty, the standard comparison operators cannot properly distinguish objective values.
Therefore, they should be substituted by operators that incorporate the existing uncertainty into the comparison. 
This section investigates various comparison operators that are based on \emph{probabilistic dominance} between instances of uncertain objectives, with a special focus on two novel operators.

Probabilistic dominance is originally defined to describe the probability that a solution $A$ dominates another solution $B$, see~\cite{Teich01} and~\cite{Hughes01}.
This probability is calculated as the product of all probabilities that an objective from $A$ is more favorable than the same objective from $B$. 
For a maximization problem, this probability can be calculated as follows:
\begin{align}
  \Pr(A \succ B) = \prod_{i=1}^{m} \Pr\big(\obj_i(A) > \obj_i(B)\big)\enspace. \label{eq:probabilistic_dominance}
\end{align}
A threshold value can be used to determine if the resulting probability is significant enough to assume $A$ dominates $B$.
The main limitation of this approach is that Equation~\eqref{eq:probabilistic_dominance} can only be applied if all $m$ objective functions are statistically independent, which is most often not true because objective functions usually have common decision variables in their inputs.
To overcome this limitation, we proposed in~\cite{KBT18} to calculate the probability $\Pr\big(\obj_i(A) > \obj_i(B)\big)$ for each objective $\obj_i$ separately, to distinguish $\obj_i(A)$ and $\obj_i(B)$ using a threshold value, and then to determine dominance between $A$ and $B$ according to Equation~\eqref{eq:uncertainty_optimization_dominance}.

\begin{figure}[t]
  \centering
  \input{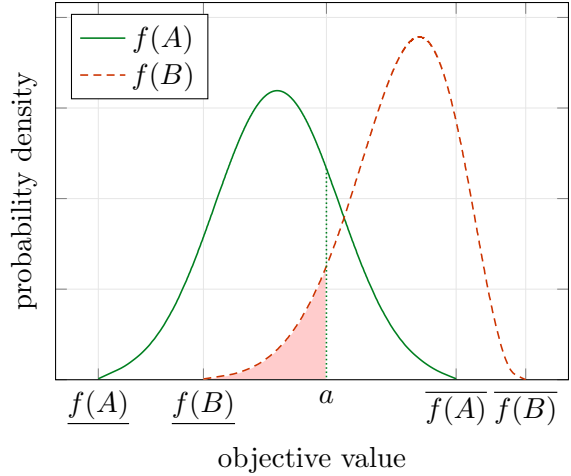}
  \caption{Example of \acp{PDF} of one uncertain objective of two candidate solutions $A$ and $B$ under comparison.}
  \vspace*{-10pt}
  \label{fig:example_distribution}
\end{figure}

For arbitrary distributions of $\obj(A)$ and $\obj(B)$, as closed-form \acp{PDF} or sample data, the probability that $\obj(A)$ is greater than $\obj(B)$ can be calculated as follows\footnote{Since the approach is applied to each objective function $\obj_i(\cdot)$ separately, from here on this notation is simply written as $\obj(\cdot).$} (see also~\cite{SI09}):
\begin{align}
\Pr\big(\obj(A) > \obj(B)\big)=
\int_{\underline{\obj(A)}}^{\overline{\obj(A)}}\Pr\big(\obj(A)=a\big)\Pr\big(\obj(B)<a\big)\mathrm{d}a\enspace. \label{eq:probabilistic_dominance_i}
\end{align}
where $\underline{\obj(A)}$ and $\overline{\obj(A)}$ denote the lower and upper bounds on $\obj(A)$, respectively.
While $\Pr\big(\obj(A)=a\big)$ is obtained from the \ac{PDF} of $\obj(A)$, the second probability $\Pr\big(\obj(B)<a\big)$ can be calculated as follows: 
\begin{equation}
  \Pr\big(\obj(B)<a\big)=\int_{\underline{\obj(B)}}^{a}\Pr(\obj(B)=b)\mathrm{d}b\enspace. \label{eq:probabilistic_dominance_ii}
\end{equation}
Figure~\ref{fig:example_distribution} shows an example of \acp{PDF} for $\obj(A)$ and $\obj(B)$.  
The filled area under the curve of $\obj(B)$ in this figure amounts to the probability $\Pr\big(\obj(B)<a\big)$ for a given value of $a$.
The exact calculation of these integrals is tedious if the \acp{PDF} of $\obj(A)$ and $\obj(B)$ are not available or if they do not follow instances of Uniform and Gaussian distributions. 
The following subsections first describe various approaches for the estimation of $\Pr\big(\obj(A) > \obj(B)\big)$ as well as other robust comparison operators.
Then, these are evaluated with respect to estimation error and execution time.

\subsection{Robust Comparison Operators}\label{sec:prob_dom_operators}
This subsection describes various comparison operators reviewed in Section~\ref{sec:relatedwork} along with two novel approaches to enable distinguishing instances of an uncertain objective. 
These operators require the distribution of uncertain objective values $\obj(A)$ and $\obj(B)$ given as \acp{PDF} or sample data.
If the \acp{PDF} are known, different statistics of the distributions can be obtained, including the expected values ($\E{\obj(\cdot)}$), variances ($\Var{\obj(\cdot)}$), standard deviations
($\sigma[\obj(\cdot)]$), $p$-th quantiles ($\q{p}{\obj(\cdot)}$) and values of the \acp{CDF} ($\CDF{a}{\obj(\cdot)}=\Pr\big(\obj(\cdot) \leq a\big)$).
If the \ac{PDF} of an uncertain objective value is not available, it can be represented by a population of samples where each sample is an outcome of evaluating the corresponding objective function.
Given a sequence $S$ of $N$ independent samples $(s_1, s_2, \ldots, s_N)$, the sample mean and the (unbiased) sample variance and standard deviation can be estimated as follows: 
\begin{equation}
    \EEst{S}=\frac{1}{N}\sum_{i=1}^{N}s_i\enspace,
    \label{eq:sample_mean}
\end{equation}
\begin{equation}
    \VarEst{S}=\frac{1}{N-1}\sum_{i=1}^{N}\big(s_i-\EEst{S}\big)^2\enspace,
\end{equation}
\begin{equation}
    \sigmaEst{S}=\sqrt{\VarEst{S}}\enspace.
\end{equation}
Also, the $p$-th quantile of a population can be calculated by the inverse empirical distribution function that traverses the population in the ascending order of samples and returns the very first sample after the $p\cdot N$ smallest
samples.
More details on the aforementioned statistics can be found in \cite{durrett2010probability}.

\subsubsection{Pairwise Comparison ($\APW$)}
The most straightforward approach to estimate the probability $\Pr\big(\obj(A) > \obj(B)\big)$
is to generate $N$ samples $(a_1,a_2,\ldots,a_N)$ from the distribution of $\obj(A)$ and
$(b_1,b_2,\ldots,b_N)$ according to the distribution of $\obj(B)$, and to calculate the proportion of pairs
$(a_i,b_i)$ where $a_i>b_i$:
\begin{equation}
  \Pr\big(\obj(A)>\obj(B)\big)\approx\frac{1}{\,N\,}\,\big\vert\{i\mid a_i>b_i\}\big\vert\enspace. \label{eq:pairwise}
\end{equation}
The absolute approximation error of this comparison operator can be determined as follows:
\begin{equation}
  \ePW = \Big\vert \Pr\big(\obj(A)>\obj(B)\big) - \frac{1}{\,N\,}\,\big\vert\{i\mid a_i>b_i\}\big\vert\Big\vert\enspace. \label{eq:pairwise_error}
\end{equation}
The probability that this error is larger than a constant tolerance $0<\delta <1$ tends exponentially to zero for large values of $N$ according to Chernoff bounds:
\begin{equation}
  \Pr\left(\ePW>\delta\right) \leq \mathrm{e}^{-c\cdot\delta^2\cdot N}\enspace,
  \label{eq:pairwiseError}
\end{equation}
where $c$ is a positive constant.

The time complexity of the comparison is of the order $O(N)$. 
Note that if the evaluation of objective functions is time-consuming, it is better to maintain a population of samples for each objective of each candidate solution throughout the optimization.
Otherwise, new samples can be generated whenever a comparison takes place.

\subsubsection{Uniform Approximation ($\AUniform$)~\cite{Teich01}}
Another approach to estimate $\Pr\big(\obj(A) > \obj(B)\big)$ is based on the assumption that $\obj(A)$ and $\obj(B)$ can be approximated by uniform distributions $\UniformNoPar(\underline{\obj(A)}, \overline{\obj(A)})$ and $\UniformNoPar(\underline{\obj(B)}, \overline{\obj(B)})$, respectively.
This approach is denoted by $\AUniform^1$.
Assuming uniform distributions, $\Pr\big(\obj(A) > \obj(B)\big)$ can be calculated as follows according to the law of total probability\footnote{For cases $\overline{\obj(A)}\leq \underline{\obj(B)}$ and $\overline{\obj(B)}\leq \underline{\obj(A)}$ there is no need for applying Equation~\eqref{eq:dominanceUniform} as the probability $\Pr\big(\obj(A)>\obj(B)\big)$ amounts to 0 and 1, respectively.}: 
\begin{align}
\Pr\big(\obj(A)>\obj(B)\big) =\Pr\big(\obj(A)>\obj(B) \wedge \obj(A) \leq \overline{\obj(B)}\,\big) 
+ \Pr\big(\obj(A)>\overline{\obj(B)}\,\big)\enspace,\label{eq:dominanceUniform}
\end{align}
where
\begin{align}
\Pr\big(\obj(A)>\obj(B) \wedge \obj(A) \leq \overline{\obj(B)}\,\big) &=
\frac{\Big(\max\{\underline{\obj(A)},\underline{\obj(B)}\}\Big)^2-\Big(\min\{\overline{\obj(A)},\overline{\obj(B)}\}\Big)^2}{2\,\Big(\overline{\obj(A)} - \underline{\obj(A)}\Big)\,\Big(\overline{\obj(B)} - \underline{\obj(B)}\Big)}  \nonumber \\ 
&+\frac{\Big(\max\{\underline{\obj(A)},\underline{\obj(B)}\}-\min\{\overline{\obj(A)},\overline{\obj(B)}\}\Big)\,\underline{\obj(B)}}{\Big(\overline{\obj(A)} - \underline{\obj(A)}\Big)\,\Big(\overline{\obj(B)} - \underline{\obj(B)}\Big)} \enspace,\label{eq:dominanceUniformII}
\end{align}
and 
\begin{align}
\Pr\big(\obj(A)>\overline{\obj(B)}\,\big)=
\frac{\overline{\obj(B)}-\min\{\overline{\obj(A)},\overline{\obj(B)}\}}{\overline{\obj(A)} - \underline{\obj(A)}}\enspace.
\end{align}

This comparison operator requires the distribution of an uncertain objective value to have finite bounds.
Alternatively, these lower and upper bounds can be calculated as $\E{\obj(\cdot)}-\sqrt{3\Var{\obj(\cdot)}}$ and $\E{\obj(\cdot)}+\sqrt{3\Var{\obj(\cdot)}}$, respectively, as the variance of a uniformly distributed
random variable $\obj(\cdot)\sim\UniformNoPar\big(\,\underline{\obj(\cdot)},\overline{\obj(\cdot)}\,\big)$ equals $\big(\,\overline{\obj(\cdot)}-\underline{\obj(\cdot)}\,\big)^2/12$.
This approach is denoted as $\AUniform^2$.
Note that treating an arbitrary distribution as a uniform distribution can impose a significant estimation error in the calculation of $\Pr\big(\obj(A)>\obj(B)\big)$, but if the distributions are actually uniform then both versions accurately represent the given distributions.
If the lower and upper bounds or alternatively the expectation and variance have to be derived from a population of samples, the time needed for sample generation, \ie objective function evaluation, which has the order $O(N)$ can increase the execution time substantially.
Otherwise, the execution time is solely spent for performing the comparison which is of order $O(1)$.

\subsubsection{Gaussian Approximation ($\AGauss$)~\cite{Hughes01}}
This approach assumes that the uncertain objective values $\obj(A)$ and $\obj(B)$ follow instances of Gaussian distributions, denoted as
$\Gauss{\E{\obj(A)}}{\Var{\obj(A)}}$ and $\Gauss{\E{\obj(B)}}{\Var{\obj(B)}}$, respectively.
Thus, the probability of $\obj(A)$ being greater than $\obj(B)$ can be estimated as follows:
\begin{align}
    \Pr\big(\obj(A)>\obj(B)\big)&=\Pr(\obj(B)-\obj(A)<0) \nonumber \\
    &=\frac{1}{2}\left( 1+\erf\left( \frac{\E{\obj(A)}-\E{\obj(B)}}{\sqrt{2\,(\Var{\obj(A)}+\Var{\obj(B)}\,)}} \right) \right)\,.\label{eq:dominanceGauss}
\end{align}
Here, $\big(\obj(B)-\obj(A)\big)\sim\GaussNoPar\big(\E{\obj(A)}-\E{\obj(B)}, \Var{\obj(A)}$ $+\Var{\obj(B)}\big)$, and $\erf$ is the error function
\begin{align}
\erf(x)=\frac{2}{\sqrt{\pi}}\int_{0}^{x}\mathrm{e}^{-t^2}\mathrm{d}t\enspace.\label{eq:erf_gauss}
\end{align}
Different approximations of the error function can be used in this comparison operator. 
Some useful and fast approximations can be found in~\cite{AS64}.

If the distribution of an uncertain objective value is not available, its expected value and variance can be derived by generating a population of samples and applying the respective estimators to this population.
This increases the time complexity from $O(1)$ to $O(N)$.
Note that this comparison operator may be subject to a noticeable approximation error since the actual distributions might be non-Gaussian.

\subsubsection{Histogram Approximation ($\AHistogram{\omega}$)~\cite{KBT18}}
This approach is based on representing uncertain objective values using histogram-based distributions. 
Such a distribution partitions the actual \ac{PDF} into intervals and considers a uniform distribution within each interval.
A column is considered in each interval such that the area covered by the column equals the probability of the actual distribution within the corresponding interval.
Given a uniform distribution within each interval, histogram-based distributions are linear combinations --- or more precisely affine combinations --- of uniform distributions.
Therefore, $\Pr\big(\obj(A)>\obj(B)\big)$ can be calculated by an affine combination of Equation~\eqref{eq:dominanceUniform}. 
To reduce computational overhead, we proposed in~\cite{KBT18} to use histograms with a fixed interval width $\omega$.
Also, the columns are aligned to the intervals $I_{\omega,k}:=[k\cdot \omega,(k+1)\cdot \omega[$ for any integer $k$ such that the columns of different histograms either perfectly overlap or are disjoint.

Given histogram-based distributions of uncertain objective values $\obj(A)$ and $\obj(B)$,
\begin{align}
    \Pr\big(\obj(A)>\obj(B)\big) =
    \sum_{k}\bigg(&\frac{1}{2}
      \Pr\big(\obj(A)\in I_{\omega,k}\big)\cdot \Pr\big(\obj(B)\in I_{\omega,k}\big) \nonumber \\
    &+\Pr\big(\obj(A)\in I_{\omega,k}\big)\cdot\Pr\big(\obj(B)<k\cdot \omega\big)\bigg)\enspace ,
  \label{eq:dominanceHistogram}
\end{align}
where  
\begin{align}
\Big\lfloor \underline{\obj(A)}/\omega\Big\rfloor \leq k \leq \Big\lceil \overline{\obj(A)}/\omega\Big\rceil\enspace . 
\label{eq:histogram_column_index_range}
\end{align}
The values for $\Pr\big(\obj(\cdot)\in I_{\omega,k}\big)$ can be accurately evaluated if the actual \ac{PDF} is available or they can be estimated as the proportion of samples that lie within $I_{\omega,k}$.
Therefore, a histogram-based distribution can be constructed by calculating $\big\lceil\overline{\obj(\cdot)}/\omega\big\rceil - \big\lceil\underline{\obj(\cdot)}/\omega\big\rceil + 1$ probabilities, one for each interval, or by generating $N$ samples and calculating this proportion for each interval.
Note that the number of intervals can be different when sampling is used, especially for small values of $N$, because not all potential intervals might be filled with one or more samples.

The probabilities $\Pr\big(\obj(\cdot)\in I_{\omega,k}\big)$ and $\Pr\big(\obj(\cdot)<\omega\cdot k\big)$ can be pre-calculated without increasing the time complexity of histogram preparation.
The calculation of $\Pr\big(\obj(A)>\obj(B)\big)$ according to the Equation~\eqref{eq:dominanceHistogram} has then a time complexity of $O\big(\big\lceil( \overline{\obj(A)}-\underline{\obj(A)})/\omega\big\rceil\big)$.
The error of calculating this probability can be bounded by the sum
\begin{align}
  \frac{1}{2}\cdot\sum_{k}
  \Pr\left(\obj(A)\in I_{\omega,k}\right)\cdot \Pr\left(\obj(B)\in I_{\omega,k}\right)\enspace,
\end{align}
where $k$ ranges according to Equation~\eqref{eq:histogram_column_index_range}.
This error is due to the loss of information on the exact distribution within the intervals.
Note that if a column from the histogram of $\obj(A)$ perfectly overlaps with a column from the histogram of $\obj(B)$,
then the values in each of the columns are assumed to be greater than the values from the other column half the time --- resulting in a probability of $0.5$.
While depending on the distribution of uncertain objective values within the shared interval, one distribution can always offer greater values which yields a probability of $1$.
This can be the case especially when $\omega$ is sufficiently large.
Furthermore, if the probabilities in Equation~\eqref{eq:dominanceHistogram} are estimated by samples, the comparison is subject to an additional error that is due to the difference between the proportion of samples within an interval and the actual probability to have a value within that interval.
The probability that this additional error is equal to or greater than $\delta$ is bounded by $e^{-c\cdot\delta^2\cdot N}$ for large values of $N$ and $0<\delta<1$, where $c$ is a positive constant and $N$ is the number of used samples.
This result can be obtained by application of the Dvoretzky-Kiefer-Wolfowitz inequality~\cite{DKW56} comparing the \acp{CDF} of the two histograms constructed according to the actual distribution and through sampling.

The execution time and accuracy of this approach significantly depend on the chosen value of $\omega$ such that shrinking $\omega$ reduces the approximation error but increases $\big\lceil( \overline{\obj(A)}-\underline{\obj(A)})/\omega\big\rceil$ which in turn prolongs the comparison, while expanding $\omega$ speeds up the comparison at the cost of increased approximation error. 
In most cases, a rough idea on how the objective values are distributed can be established, and thus, a good choice for $\omega$ can be made, at least after some prior experiments.

\subsubsection{Proposed Empirical Distribution-based Approach ($\AEmpirical$)}
The distribution of uncertain objective values often cannot be fitted to a closed-form \ac{PDF}, and therefore, must be represented by a population of samples. 
To achieve a good tradeoff between the accuracy and execution time of the comparison, we propose here an approach that is based on the empirical distribution of uncertain objective values. 
Given the population of samples $S = (s_1, s_2, \ldots, s_N)$, a random variable $X$ which is distributed according to the corresponding empirical distribution has a \ac{CDF}
\begin{equation}
\CDF{y}{X}=\Pr(X\leq y)=\frac{1}{N}\big\vert\{i\mid s_i\leq
y\}\big\vert\enspace,
\end{equation}
which amounts to the proportion of samples being smaller than or equal to $y$.

Let $\obj(A)$ and $\obj(B)$ follow two empirical distributions that are specified by populations of samples $(a_1, a_2, \ldots, a_N)$ and $(b_1, b_2, \ldots, b_M)$, respectively. 
A naive approach to compare these uncertain objective values would then be
\begin{equation}
  \Pr(f(A) > f(B))=\frac{\vert\{(i,j)\mid a_i>b_j\}\vert}{N\cdot M}\enspace.
  \label{eq:empirical}
\end{equation}
Unlike the pairwise comparison $\APW$, each sample in the first population is not only compared to the corresponding sample but also to all other samples in the second population.
The desired probability is then estimated as the proportion of pairs wherein the sample from the first population is greater than the sample from the second population.

The time complexity of this approach is of order $O(N\cdot M)$, or $O(N^2)$ if $M=N$, which is usually the case.
To reduce this complexity, we propose to first sort each population of samples in ascending order and then to apply the function shown in Algorithm~\ref{alg:compareEmpirical}.
This function receives sorted lists $(a_1,\ldots,a_N)$ and $(b_1,\ldots,b_M)$ that respectively represent the uncertain objective values $\obj(A)$ and $\obj(B)$, and returns the probability $\Pr\big(\obj(A)>\obj(B)\big)$.
It uses two indices $i$ and $j$ to traverse these lists in ascending order.
It stores the proportion of pairs wherein the sample $a_i$ is greater than the sample $b_j$ in a variable named $num\_pairs$.  
For each $a_i$, it adds $j-1$ to $num\_pairs$ when $b_j$ is the first element in its list that is not smaller than $a_i$.
Since each list is traversed only once, the comparison is performed in linear time $O(N+M)$, or simply $O(N)$ if $M=N$.
Also, the condition and the body of the while loop are evaluated at most $N+M$ and $M$ times, respectively.
Note that generating and sorting the samples have the time complexities of respectively $O(N+M)$ and $O(N\log N + M\log M)$.

\begin{algorithm}[t]
  \SetKwInOut{Input}{Input}%
  \SetKw{KwTo}{to}%
  \SetKw{KwAnd}{and}%
  \caption{Comparing sorted lists of samples}
  \label{alg:compareEmpirical}
  \Input{Sorted lists $(a_1,\ldots,a_N)$ and $(b_1,\ldots,b_M)$}
  $j:=1$\;
  ${\it num\_pairs}:=0$\;
  \For{$i:=1$ \KwTo $N$}
  {
    \While{$j<=M$ \KwAnd $a_{i}>b_{j}$}
    {
      $j:=j+1$\;
    }
    ${\it num\_pairs}:={\it num\_pairs}+j-1$\;
  }
  \Return ${\it num\_pairs}/(N\cdot M)$\;
\end{algorithm}

\begin{figure}
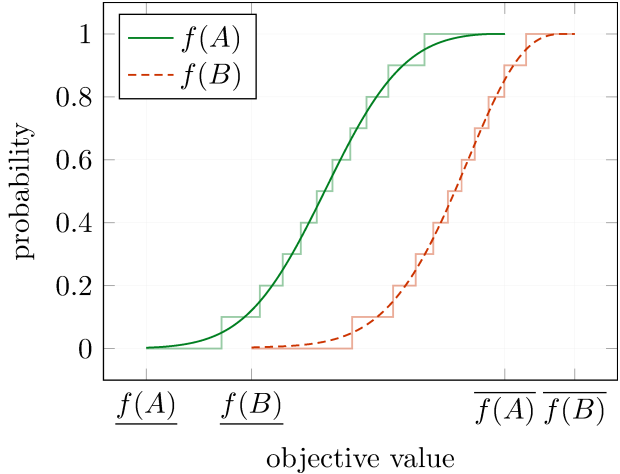

  \centering
  \cdfExample
  \caption{Exact and approximated \acp{CDF} for the distributions shown in Figure~\ref{fig:example_distribution}: the approximation uses 10 quantiles.}
  \label{fig:example_cdf}
\end{figure}

If the \acp{PDF} of $\obj(A)$ and $\obj(B)$ are available, the proposed approach can be extended to derive $N$ quantiles that partition the actual distribution into intervals of equal probabilities. 
This enables to construct empirical distributions that achieve better approximations of the actual distributions.
Figure~\ref{fig:example_cdf} shows the exact \acp{CDF} of the distributions shown in Figure~\ref{fig:example_distribution} as well as approximations of these \acp{CDF} using 10 quantiles.
Using the quantiles such that $a_i=\q{\frac{2i-1}{2N}}{}$, the difference between the \acp{CDF} of the empirical and the actual distributions are restricted to $\frac{1}{2N}$.
Considering this difference for both $\obj(A)$ and $\obj(B)$ results in a maximum estimation error of $1/N$ for the calculation of $\Pr\big(\obj(A)>\obj(B)\big)$.
Let $X_A$ and $X_B$ be random variables distributed according to the empirical distributions of $\obj(A)$ and $\obj(B)$, respectively.
Then
\begin{align}
  \Pr\big(\obj(A)>\obj(B)\big)
  &=\int_{\underline{\obj(A)}}^{\overline{\obj(A)}}\Pr\big(\obj(A)=a\big)\Pr\big(\obj(B)<a\big)\mathrm{d}a \nonumber \\
  &=\int_{\underline{\obj(A)}}^{\overline{\obj(A)}}\Pr\big(\obj(A)=a\big)\big(\Pr(X_B<a) + \varepsilon(a)\big)\mathrm{d}a \nonumber \\
  &=\varepsilon+\int_{\underline{\obj(A)}}^{\overline{\obj(A)}}\Pr(\obj(A)=a)\Pr(X_B<a)\mathrm{d}a \nonumber \\
  &=\varepsilon+\int_{\underline{\obj(B)}}^{\overline{\obj(B)}}\Pr(\obj(A)\geq b)\Pr(X_B=b)\mathrm{d}b \nonumber \\
  &=\varepsilon+\int_{\underline{\obj(B)}}^{\overline{\obj(B)}}(\Pr(X_A\geq b)+ \varepsilon'(b))\Pr(X_B=b)\mathrm{d}b \nonumber \\
  &=\varepsilon+\varepsilon'+\int_{\underline{\obj(B)}}^{\overline{\obj(B)}}\Pr(X_A\geq b)\Pr(X_B=b)\mathrm{d}b \nonumber \\
  &=\varepsilon+\varepsilon'+\Pr(X_A>X_B)\enspace,  \label{eq:errorbound}
\end{align}
where $\varepsilon(x),\varepsilon,\varepsilon'(x),\varepsilon'\in[-1/(2N),1/(2N)]$ for all $x \in [\min\{\underline{\obj(A)},\underline{\obj(B)}\}, \max\{\overline{\obj(A)},\overline{\obj(B)}\}]$.
Note that the quantiles are already sorted in the ascending order, and the time needed for deriving $N$ quantiles is of order $O(N)$.

Similar to the pairwise comparison $\APW$, the error of the empirical distribution can be bounded.
Let $X$ be a random variable that follows the empirical distribution of $\obj(\cdot)$ produced from a population of $N$ samples.
Then, according to the Dvoretzky-Kiefer-Wolfowitz inequality~\cite{DKW56}
\begin{equation}
\Pr\left(\sup_{x\in\R}\big\vert\CDF{x}{\obj(\cdot)}-\CDF{x}{X}\big\vert>\delta\right)\leq \mathrm{e}^{-c\cdot\delta^2\cdot N}
\label{eq:empiricalerrorprobability}
\end{equation}
for large values of $N$ and $0<\delta<1$, where $c$ is a positive constant. 
The constants used here are worse than those used for $\APW$ as the result for the pairwise comparison of random variables $\obj(A)$ and $\obj(B)$ can be seen as the value of the \ac{CDF} of their difference at the point zero $\CDF{0}{\obj(B)-\obj(A)}$. 
Here, instead, the maximal error among all input values is bounded.
The actual error of estimating $\Pr\big(\obj(A)>\obj(B)\big)$ is then bounded by the sum of the two errors introduced by the empirical distributions of $\obj(A)$ and $\obj(B)$. 
Nevertheless, the experiments in Section~\ref{sec:evaluation_comparison_operators} imply that this estimation is much more accurate.

\subsubsection{Proposed Reduced Empirical Distribution-based Approach ($\AReduced$)}
The proposed empirical distribution-based comparison operator can be further improved for the case where the \acp{PDF} of the uncertain objective values are unknown and can only be approximated through sampling.
This approach is similar to the histogram approximation ($\AHistogram{\omega}$) with the exception that the partitioning is performed evenly in the probability domain instead of the domain of the objective values.
It is used only if at least one of the uncertain objective values is given as a population of samples rather than a \ac{PDF}.
Otherwise, we use quantiles as described for the empirical distribution ($\AEmpirical$) to provide a better approximation.

If we have a sorted list of samples $(s_1,\ldots,s_N)$, then we will use only
$N'=\Theta(\sqrt{N})$ data points $(y_1,\ldots,y_{N'})$, where $y_i=s_{\lceil (i-1/2)\cdot
N/N'\rceil}$ to produce the empirical distribution.
In our experiments, we will use
exactly $\lceil \sqrt{N}\,\rceil$ data points.
One could use pivot-based sorting algorithms like quicksort for sorting the
samples $(s_1,\ldots,s_N)$ and stop sorting if none of the needed indices are
available in an interval which has to be sorted currently, but as this does
not change the (expected) complexity of $O(N\log N)$ for sorting, we do not
elaborate on this improvement.
Therefore, for initialization, we still have the same complexity as for $\AEmpirical$
with $N$ sample generations plus $O(N \log N)$ for sorting. The comparison of
two uncertain objective values is then much faster.  Let $(a_1,\ldots,a_{N'})$,
$(b_1,\ldots,b_{M'})$ be the reduced list of sorted samples from uncertain objective values $\obj(A)$ and $\obj(B)$, where $N'=\Theta(\sqrt{N})$ and $M'=\Theta(\sqrt{M})$,
then we can use again Algorithm~\ref{alg:compareEmpirical} to obtain the
complexity $O(\sqrt{N}+\sqrt{M})$.

The error caused by approximation with an empirical distribution using only $\sqrt{N}$ samples is then automatically in $O(1/\sqrt{N})$ which can be obtained by an equation analogous to Equation~\eqref{eq:errorbound}.
Now we analyze the situation when we pick a specific position $z$ and query the
probability of a random variable $X$ to be less than $z$. 
Let $p$ be the actual
probability. For $N$ samples the relative number of samples less than $z$ is exactly
the value of the empirical distribution at $z$. This value multiplied by $N$ has
a binomial distribution with parameter $p$ and its variance scales with ${N}$.
Consequently the variance of the empirical distribution evaluated on any
position scales with $1/N$ and the standard deviation scales with $1/\sqrt{N}$.
Therefore, the order of the error does not change if the number of data points
is reduced to only the square root of the number of evaluated samples as
suggested.

Table~\ref{tab:complexity} summarizes the analyzed time complexities of the initialization, \ie approximating distributions, and comparison for all comparison operators investigated in this section. 
It also reports if the result of the respective comparison operator converges to the exact probability $\Pr\big(\obj(A)>\obj(B)\big)$. 
Note that the initialization is applicable only if the actual \ac{PDF} of at least one objective function is not available. 
Also, in the case of the proposed empirical distribution-based approach ($\AEmpirical$), $N$ denotes either the number of quantiles or the number of samples, depending on whether the actual \acp{PDF} are available or not. 

\begin{table}
  \caption{Time complexity for initialization and comparison of the investigated comparison operators, and the information whether their approximation error tends to zero if the number
of samples $N$ is increased.}
\vspace*{-10pt}
  \begin{center}
  \renewcommand{\arraystretch}{1.4}
  \small
  \begin{tabular}{ l | c | c | c }
         &initialization        &comparison         &error$\rightarrow 0$\\
         \hline
         $\APW$                 & ---               & $O(N)$ & \cmark\\
         $\AUniform^{1/2}$      & $O(N)$            & $O(1)$ & \xmark\\
         $\AGauss$              & $O(N)$            & $O(1)$ & \xmark\\
         $\AHistogram{\omega}$  & $O\big(N + \big\lceil\frac{\overline{\obj(\cdot)}-\underline{\obj(\cdot)}}{\omega}\big\rceil\big)$ & $O(\big\lceil\frac{\overline{\obj(\cdot)}-\underline{\obj(\cdot)}}{\omega}\big\rceil\big)$ & \xmark\\
         $\AEmpirical$          & $O(N\log N)$      & $O(N)$ & \cmark\\
         $\AReduced$            & $O(N\log N)$      & $O(\sqrt{N})$ & \cmark
  \end{tabular}
  \label{tab:complexity}
\end{center}
\vspace*{-5pt}
\end{table}

\begin{figure}
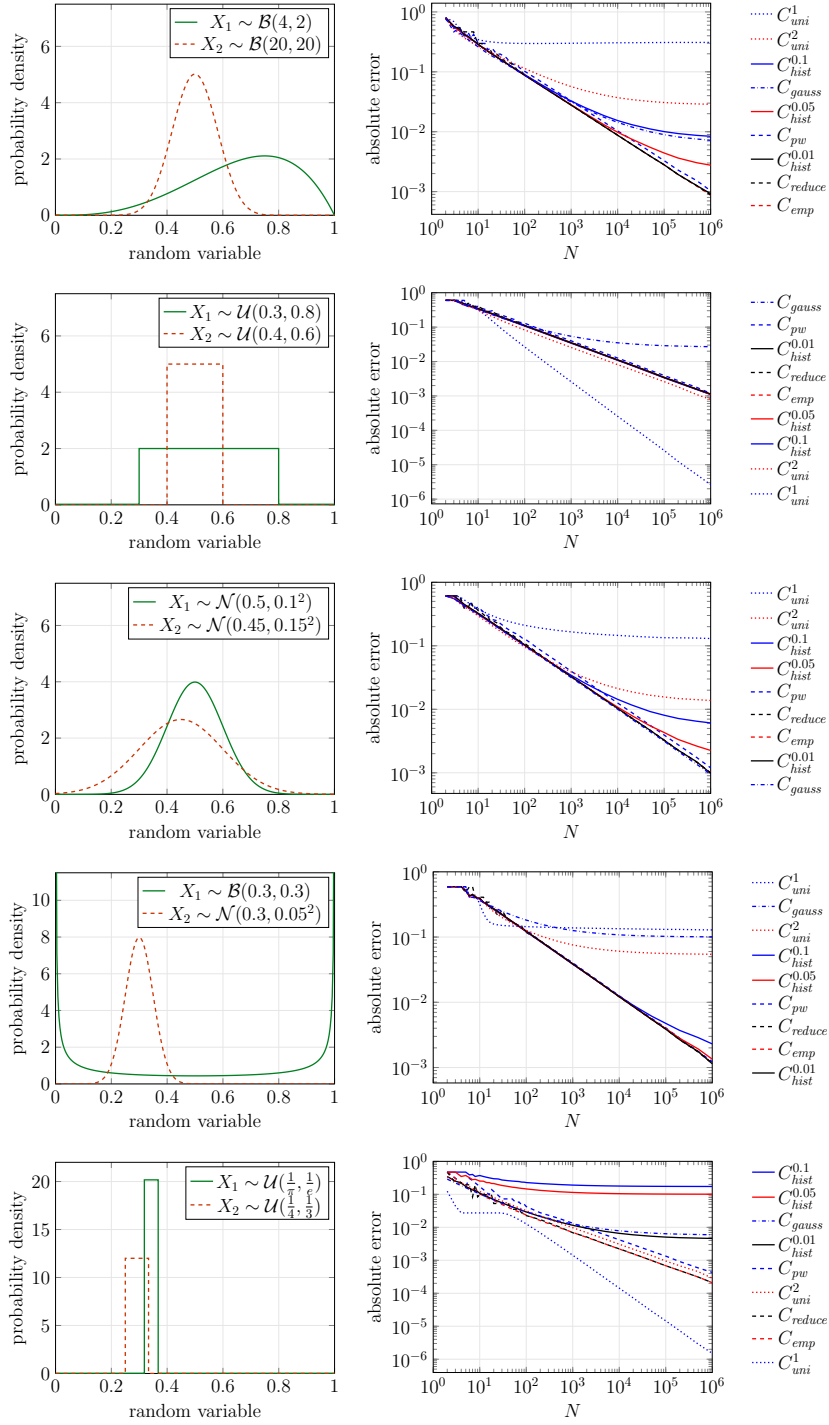

  \begin{center}
	  \begin{tabular}{@{}c@{}}
    \hspace*{-2mm}
    \distI
    \hspace*{-2mm}
    \compQualityI
    \hspace*{-2mm}
    \compQualityLegendI
    \\[0.5\baselineskip]
    \hspace*{-2mm}
    \distII
    \hspace*{-2mm}
    \compQualityII
    \hspace*{-2mm}
    \compQualityLegendII
    \\[0.5\baselineskip]
    \hspace*{-2mm}
    \distIII
    \hspace*{-2mm}
    \compQualityIII
    \hspace*{-2mm}
    \compQualityLegendIII
    \\[0.5\baselineskip]
    \hspace*{-2mm}
    \distIV
    \hspace*{-2mm}
    \compQualityIV
    \hspace*{-2mm}
    \compQualityLegendIV
    \\[0.5\baselineskip]
    \hspace*{-2mm}
    \distV
    \hspace*{-2mm}
    \compQualityV
    \hspace*{-2mm}
    \compQualityLegendV
  \end{tabular}
\end{center}
\vspace*{-15pt}
    \caption{Left: Five scenarios of two random variables with different distributions introduced for evaluating different comparison operators. 
    Right: The absolute error of comparison operators for each scenario and for different numbers $N$ of samples.
    The comparison operators in the legend are ranked according to their absolute error bound with the maximal number of samples $N=10^6$.}
  \label{fig:scenarios}
\end{figure}

In the following, we evaluate the investigated comparison operators in terms of approximation error and execution time. 
For the evaluation, five scenarios of two random variables $(X_1, X_2)$ with different distributions are selected, see the left column of Figure~\ref{fig:scenarios}.
These scenarios combine instances of various distributions including uniform $\UniformNoPar(\underline{X}, \overline{X})$, Gaussian $\Gauss{\E{X}}{\Var{X}}$ and beta 
$\BetaDistribution{\alpha}{\beta}$ distributions.
The first four scenarios are adopted from~\cite{KBT18}.
These scenarios vary in their statistical properties and pose different challenges to the comparison operators. 

\subsection{Approximation Error Analysis of Comparison Operators}\label{sec:evaluation_comparison_operators}
The right part of Figure~\ref{fig:scenarios} displays the quality of
different comparison operators for each scenario.  For each scenario and each
comparison operator one can see the development of the absolute error -
absolute difference of the evaluated value and the correct value - while the
number of samples is increased.
To be more precise the presented graphs represent an error bound such that
$99\%$ of comparisons by the respective comparison operator comply with that
error bound. This limitation is necessary because even with large numbers of
samples all samples of one random variable can be smaller than all samples of
the other random variable even if the first random variable dominates the
second random variable with probability greater than $50\%$. In such cases all
comparison operators will fail badly but fortunately, such an event will most
likely not happen.
To obtain additional information we also provide in
Table~\ref{tab:differences} the differences between the actual dominance values
and the evaluated dominance values by each comparison operator if infinitely
many samples would have been used. $\APW$, $\AEmpirical$ and $\AReduced$ are
omitted as they always converge to the actual dominance values.

\begin{table}
  \caption{For the scenarios in Figure~\ref{fig:scenarios}, the actual values $\Pr(X_1>X_2)$
	and the values calculated by each comparison operator are reported. Additionally, the respective
difference to the actual values if the number of samples tends to infinity is shown.}
\vspace*{-10pt}
\begin{center}
\small
${\arraycolsep=3pt
  \begin{array}{c| l | l | l | l | l}
    & 1 & 2 & 3 & 4 & 5\\
    \hline
	\Pr(X_1\!\!>\!\!X_2) &   0.7978  &     0.6000 &  0.6092 &  0.5923 &     0.9727\\
    \hline
    \AUniform^1      & \begin{array}{@{}l@{}} 0.5000 \\   -0.2978\end{array}
                     & \begin{array}{@{}l@{}} 0.6000 \\\pm 0.0000\end{array}
                     & \begin{array}{@{}l@{}} 0.5000 \\   -0.1092\end{array}
                     & \begin{array}{@{}l@{}} 0.5000 \\   -0.0923\end{array}
                     & \begin{array}{@{}l@{}} 0.9727 \\\pm 0.0000\end{array} \\
\hline
    \AUniform^2      & \begin{array}{@{}l@{}} 0.7700 \\   -0.0278\end{array}
                     & \begin{array}{@{}l@{}} 0.6000 \\\pm 0.0000\end{array}
                     & \begin{array}{@{}l@{}} 0.5962 \\   -0.0130\end{array}
                     & \begin{array}{@{}l@{}} 0.6461 \\   +0.0538\end{array}
                     & \begin{array}{@{}l@{}} 0.9727 \\\pm 0.0000\end{array} \\
\hline
    \AGauss          & \begin{array}{@{}l@{}} 0.8042 \\   +0.0064\end{array}
                     & \begin{array}{@{}l@{}} 0.6261 \\   +0.0261\end{array}
                     & \begin{array}{@{}l@{}} 0.6092 \\\pm 0.0000\end{array}
                     & \begin{array}{@{}l@{}} 0.6922 \\   +0.0999\end{array}
                     & \begin{array}{@{}l@{}} 0.9669 \\   -0.0058\end{array} \\
\hline
    \AHistogram{0.1} & \begin{array}{@{}l@{}} 0.7902 \\   -0.0076\end{array}
                     & \begin{array}{@{}l@{}} 0.6000 \\\pm 0.0000\end{array}
                     & \begin{array}{@{}l@{}} 0.6040 \\   -0.0052\end{array}
                     & \begin{array}{@{}l@{}} 0.5936 \\   +0.0013\end{array}
                     & \begin{array}{@{}l@{}} 0.8000 \\   -0.1727\end{array} \\
\hline
    \AHistogram{0.05}& \begin{array}{@{}l@{}} 0.7959 \\   -0.0019 \end{array}
                     & \begin{array}{@{}l@{}} 0.6000 \\\pm 0.0000\end{array}
                     & \begin{array}{@{}l@{}} 0.6079 \\   -0.0013 \end{array}
                     & \begin{array}{@{}l@{}} 0.5926 \\   +0.0003\end{array}
                     & \begin{array}{@{}l@{}} 0.8721 \\   -0.1006 \end{array} \\
\hline
    \AHistogram{0.01}& \begin{array}{@{}l@{}} 0.7977 \\   -0.0001\end{array}
                     & \begin{array}{@{}l@{}} 0.6000 \\\pm 0.0000\end{array}
                     & \begin{array}{@{}l@{}} 0.6092 \\   -0.00005\!\!\end{array}
                     & \begin{array}{@{}l@{}} 0.5923 \\   +0.00001\!\!\end{array}
                     & \begin{array}{@{}l@{}} 0.9683 \\   -0.0044\end{array} \\
  \end{array}
}$
\end{center}
\vspace*{-5pt}
  \label{tab:differences}
\end{table}

These results visualize that the comparison operators which assume uniform or
Gaussian distributions ($\AGauss$, $\AUniform$) outperform other comparison
operators if the random variables actually follow instances of the assumed distribution but they can
also have quite large errors if this is not the case.

Also the histogram based comparison operator ($\AHistogram{\omega}$) has this problem
as it assumes piecewise constant densities. Also if the positions where
densities of the histogram can change do not match the positions where the actual distributions
change, a significant error can be received. This is most explicitly tracked in
the last scenario. But if the width $\omega$ of the columns in the histograms is suitably
chosen then it leads to similar results as for the proposed comparison operators based on empirical distributions.

The remaining comparison operators ($\APW$, $\AEmpirical$ and $\AReduced$)
finally converge to the actual dominance value if the number of samples is
increased. The comparison operator using empirical distributions
($\AEmpirical$) and its reduced version ($\AReduced$) converge in all cases
faster than the pairwise comparison ($\APW$). For larger sample sizes, there is
almost no difference between $\AEmpirical$ and $\AReduced$. Only for very small
sample sizes $\AReduced$ has a larger error due to too few data points in the
reduced empirical distribution. Also for $\APW$, the variance is quite large if
the sample size is small.

Here the presented experiments confirm the advantages and disadvantages
predicted by theoretical evaluation in the previous section.

\begin{figure*}
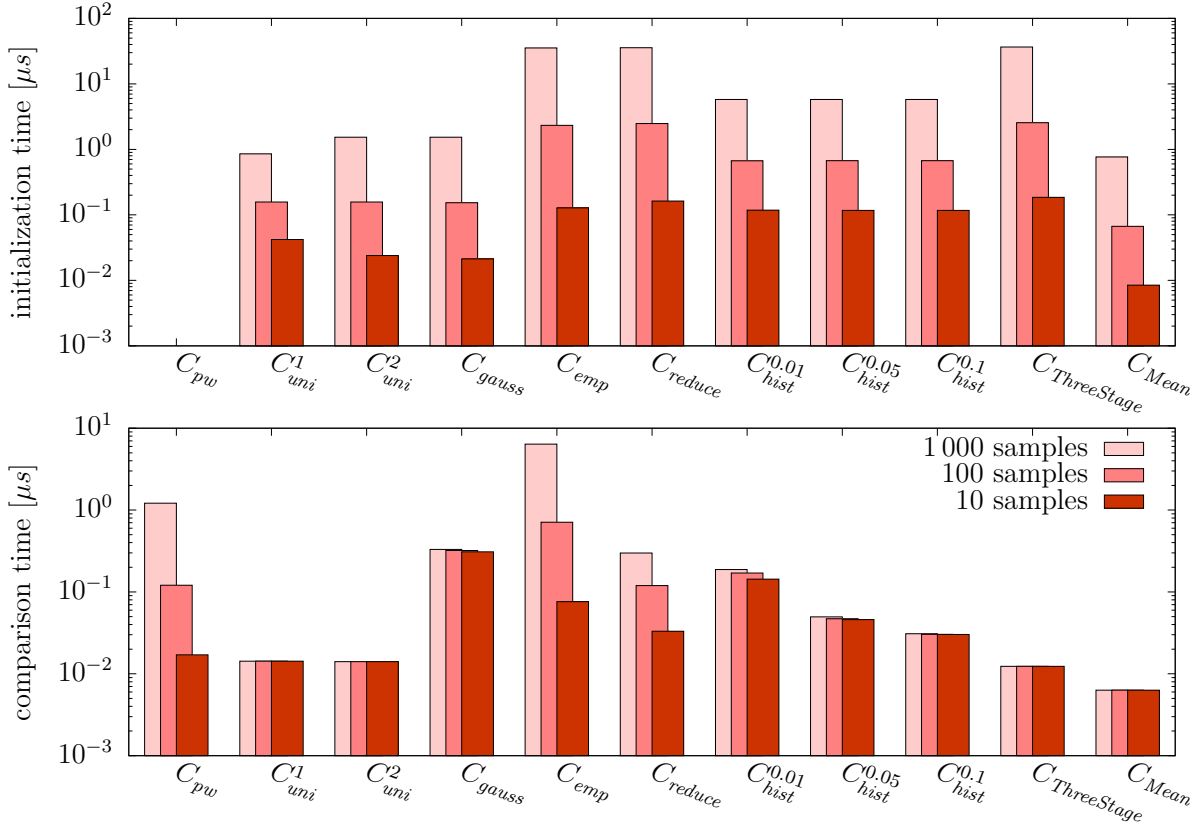

  \resizebox{\textwidth}{!}
{
  \input{figures/timingcomparatorsinitialization.tex}
}
  \resizebox{\textwidth}{!}
{
  \input{figures/timingcomparatorscomparison.tex}
}
   \caption{For each comparison operator, we display the execution times of the
    initialization of one object for comparison (once for
    each object) and the execution times of the comparison of two already initialized
  objects in $\mu s$ for $10$, $100$ and $1\,000$ samples per object
(Programming language: Java, executed on an Intel i7-4790 with an Ubuntu operating system).}
  \label{fig:timing}
\end{figure*}

\subsection{Execution Time Analysis of Comparison Operators}
In addition to the comparison operators presented in Section~\ref{sec:prob_dom_operators}, there exist other comparison operators which leave out the detour via probabilistic dominance.
These comparison operators do not calculate the dominance value and evaluate the preference of an uncertain objective value in comparison to another by different characteristics.
Two well established comparison operators of that kind are the mean-based approach~\cite{MAG12} and the three-stage comparison operator~\cite{KMGT15} which are briefly introduced in the following:

\subsubsection{Mean-based Comparison Operator ($\AMeanBased$)~\cite{MAG12}}
This comparison operator simply decides whether $\obj(A)$ or $\obj(B)$ is better by comparing their mean values.
The initialization takes linear time as the mean has to be estimated according to Equation~\eqref{eq:sample_mean} and the comparison takes only constant time.

\subsubsection{Three-Stage Comparison Operator ($\AThreeStage$)~\cite{KMGT15}}
This comparison operator performs the comparison in three stages.
In the first stage, it checks whether the domains, \ie the interval between the lower and upper bound, of the two uncertain objective values overlap.
If not, then the value in the better region is decided to be better.
If the domains do overlap, the next stages apply.
In the second stage, it checks whether the mean of one uncertain objective value is significantly better than that of the other value.
If this is true, then the means determine whether $\obj(A)$ or $\obj(B)$ is better.
Otherwise the third stage is applied where the uncertain objective values are compared with respect to their $95\%$ quantile interval, \ie the interval between the $0.025$-th and $0.975$-th quantile points.
By comparison of quantiles it is checked whether $\obj(A)$ or $\obj(B)$ shows a considerably smaller deviation and delivers a more robust objective value.
If the lengths of these quantile intervals are very similar, it is decided that $\obj(A)$ and $\obj(B)$ are equally good.
To prepare quantiles we have the complexity $O(N \log N)$ for sorting $N$
samples during initialization time of each object. The comparison is then
evaluated in constant time.

Now we relate these two comparison operators with the six probabilistic dominance-based comparison operators introduced in Section~\ref{sec:prob_dom_operators}.
For this purpose, we interpret the dominance values calculated by a comparison operator from Section~\ref{sec:prob_dom_operators} to obtain the answer for two solutions $A$ and $B$ whether $\obj(A)$ is greater/smaller than or as good as $\obj(B)$.
To accomplish this for two uncertain objective values $\obj(A)$ and $\obj(B)$, we can define a threshold $\gamma\geq
0.5$ to determine whether the probability that $\obj(A)$ is greater/smaller than $\obj(B)$ is significant enough to reach a conclusion.
For example, in case of a maximization problem we say that $\obj(A)$ is better than $\obj(B)$ if
$\Pr\big(\obj(A)>\obj(B)\big)>\gamma$, $\obj(B)$ is better than $\obj(A)$ if $\Pr\big(\obj(B)>\obj(A)\big)>\gamma$ and otherwise
they are equally good, see~\cite{KBT18}.
This probabilistic dominance criterion can incorporate any of the previously introduced comparison operators in Section~\ref{sec:prob_dom_operators} to enable treating uncertainty in multi-objective optimization.

All the previously introduced comparison operators (the six comparison operators introduced in Section~\ref{sec:prob_dom_operators} and the two well established comparison operators introduced in this section) have been implemented in Java and experimented on an Intel i7-4790 with an Ubuntu operating system.
The average execution times for the initialization and comparison of each comparison operator for the scenarios in Figure~\ref{fig:scenarios} are visualized in Figure~\ref{fig:timing}.
The measured results comply with the presented complexities.
The only surprise might be that $\AGauss$ has such a slow comparison time which is mainly spent for the approximation of the error function in Equation~\eqref{eq:erf_gauss}.

\section{Uncertain Multi-Objective Optimization Benchmark}\label{sec:benchmark}
This section presents a number of uncertain multi-objective optimization problems to evaluate the proposed comparison operator alongside the state-of-the-art approaches. 
These problems aim to incorporate various uncertainties in six multi-objective optimization problems of the DTLZ benchmark suite~\cite{DTLZ02}.
Each of these problems are specified by a vector of $m$ objective functions $(f_1, f_2, \dots, f_m)$,  that receive a vector of $n$ shared decision variables $x=(x_1, x_2, \dots, x_n)$ where $\forall i \in \{1, 2, \dots, n\}: 0 \leq x_i \leq 1$. 
 
Table~\ref{tab:problem} shows the proposed uncertain DTLZ (UDTLZ) problems. 
The considered uncertainties include perturbation in decision variables as well as noise and approximation error in objective functions\footnote{The uncertainty due to time-varying objective functions is not considered in the proposed uncertain problems. However, the proposed comparison operator can deal with any uncertainty that leads to probabilistic representations of objective values.
}.

The UDTLZ1 problem adds uncertainties specified by instances of Beta distribution $\BetaNoPar$ with different shape parameters onto the decision variables of the DTLZ1 problem. 
The beta distribution generates samples $\bm{u}$ in the interval $[0,1]$ which are scaled to $[0,0.001]$ and added to the expected values $x_i$.
The resulting value $\bm{x_i}$ is then bounded to 1 which is the upper bound of $x_i$.
Similarly, UDTLZ6 incorporates perturbations following instances of Gaussian distribution $\GaussNoPar$ in the decision variables of DTLZ6.
The Gaussian distributions in this problem generate samples around an expected value of $0$ with different standard deviations.
Since the added variations may be negative or positive, the resulting value is then bounded to $[0,1]$. 

\begin{table*}[p!]
  \caption{UDTLZ benchmark suite incorporating uncertainty into DTLZ multi-objective optimization problems: 
  $\mathcal{N}$, $\BetaNoPar$ and $\UniformNoPar$ represent Gaussian, Beta and discrete uniform distributions. 
  Shown in \textbf{bold} are the proposed modifications.}
  \label{tab:problem}
  \renewcommand{\arraystretch}{1.8}
   \centering
   {{
	\footnotesize
	{
  \begin{tabularx}{\textwidth}{c:l} 
    \Xhline{2\arrayrulewidth}
    problem  &\multicolumn{1}{c}{objective functions}    \\
    \hline
    \multirow{4}{*}[-5pt]{UDTLZ1}     &$f_1(x)= \frac{1}{2}x_1\,x_2\dots x_{m-1} \big(1+g(\bm{x_{m}}, \dots, \bm{x_{n}})\big) $   \\
	&$\displaystyle\underset{i=2}{\overset{m}{\forall}}f_i(x)= \frac{1}{2}\left(\prod_{j=1}^{m-i}x_j\right)(1-x_{m-i+1}) \big(1+g(\bm{x_{m}}, \dots, \bm{x_{n}})\big)$    \\ 
                                &{$g(\bm{x_{m}}, \dots, \bm{x_{n}}) = 100(n-m+1)\displaystyle\sum_{i=m}^{n}\Big({(x_i - 0.5)}^2 - \cos\big(20\pi(x_i - 0.5)\big)\Big)$} \\
                                &{where $\underset{\bm{i=1}}{\overset{\bm{n}}{\bm{\forall}}} \, \bm{x_i = \min(}x_i \bm{+ 0.001\, u_i, 1)}$ and $\bm{u_i \sim \BetaDistribution{10 + i}{2 + i}}$} \\
    \hline
    \multirow{4}{*}[-8pt]{UDTLZ2}     &$f_1(x)= \big(1+g(x_{m}, \dots, x_{n})\big) \bm{\mbox{$\cos$}(}\theta_1\bm{)}$\,$\dots$\,$\bm{\mbox{$\cos$}(}\theta_{m-1}\bm{)} \bm{+ u_1}$     \\
                                &$\displaystyle\underset{i=2}{\overset{m}{\forall}}f_i(x)= \big(1+g(x_{m}, \dots, x_{n})\big) \left(\prod_{j=1}^{m-i}\bm{\mbox{$\cos$}(}\theta_j\bm{)}\right)\bm{\mbox{$\sin$}(}\theta_{m-i+1}\bm{)} \bm{+ u_1}$    \\ 
                                &$\theta_i = \displaystyle\frac{\pi}{2} x_i, g(x_{m}, \, \dots, x_{n}) = \displaystyle\sum_{i=m}^{n}\big({(x_i - 0.5)}^2\big)$
                                 \\
                                &where\, $\bm{\mbox{$\sin$}(\theta) = \displaystyle\sum_{j=1}^{u_2}(-1)^j\frac{\theta^{2j+1}}{(2j+1)!}}, \,\, \bm{\mbox{$\cos$}(\theta) = \displaystyle\sum_{j=1}^{u_2}(-1)^j\frac{\theta^{2j}}{(2j)!}},$\\
								& $\bm{u_1 \sim \Gauss{0}{0.005^2}}, \,\, \bm{u_2 \sim \Uniform{3}{12}}$  \\
   \hline
 \multirow{2}{*}[-2pt]{UDTLZ3}      &{Same as in UDTLZ2 except that}\\
 									& $g(x_{m}, \dots, x_{n}) = 100(n-m+1)\displaystyle\sum_{i=m}^{n}\big({(x_i - 0.5)}^2 - \bm{\mbox{$\cos$}(}20\pi(x_i - 0.5)\bm{)}\big)$ and $\bm{u_2 \sim \Uniform{12}{19}}$ \\
   \hline
   \multirow{1}{*}[-2pt]{UDTLZ4}      &{Same as in UDTLZ2 except that $x$ is replaced by $x^{100}$ in $f_i$ functions as suggested in~\cite{DTLZ02}}\\
   \hline
   \multirow{2}{*}[-2pt]{UDTLZ5}      &Same as in UDTLZ2 except that \\&
   								$\displaystyle\underset{i=2}{\overset{m-1}{\forall}} \theta_i = \frac{\pi \big(1+2\, g(x_{m}, \dots, x_{n})\, x_i \big)}{4 \big(1+g(x_{m}, \dots, x_{n})\big)}$
                                and $g(x_{m}, \, \dots, x_{n}) = \displaystyle\sum_{i=m}^{n}{x_i}^{0.1}$\\
   \hline
   \multirow{5}{*}[-7pt]{UDTLZ6}      &{$\displaystyle\underset{i=1}{\overset{m-1}{\forall}} \, f_i(\bm{x_i})=\bm{x_i}, \quad f_{m}(\bm{x})= \big(1+g(\bm{x_{m}}, \dots, \bm{x_{n}})\big) \cdot h\big(f_1(\bm{x_1}), \dots, f_{m-1}(\bm{x_{m-1}}), g(\bm{x_{m}}, \dots, \bm{x_{n}})\big)$} \\
                                &{$g(\bm{x_{m}}, \dots, \bm{x_{n}})=1+\displaystyle\frac{9}{n-m+1}\sum_{i=m}^{n}{\bm{x_i}}$}	\\
							&{$h\big(f_1(\bm{x_1}), \dots, f_{m-1}(\bm{x_{m-1}}), g(\bm{x_{m}}, \dots, \bm{x_{n}}))\big) = $}\\
						  	&{$\qquad\qquad = m - \displaystyle\sum_{i=1}^{m-1}\bigg(\frac{f_i(x_i)}{1+g(\bm{x_{m}}, \dots, \bm{x_{n}})}\big(1+\sin\big(3\pi\,f_i(\bm{x_i})\big)\Big)\bigg)$} \\
                                &{where $\underset{\bm{i=1}}{\overset{\bm{n}}{\bm{\forall}}} \, \bm{x_i = \max\left(\min\left(}x_i \bm{+ u, 1\right), 0\right)}$ and $\bm{u \sim \mathcal{N}\left(0,\frac{10+i}{1000}\right)}$} \\
    \Xhline{2\arrayrulewidth}
  \end{tabularx}
  \renewcommand{\arraystretch}{1.2}
  }
}
}
\end{table*}
\FloatBarrier

UDTLZ2, UDTLZ3, UDTLZ4 and UDTLZ5 use both function noise and approximation error to add uncertainty to their corresponding DTLZ problems. 
Function noise $\bm{u_1}$ describes samples from a Gaussian distribution with an expected value of $0$ and a standard deviation of $0.005$ across all  
of these problems. 
Also, function approximation error is added to this problems by replacing each trigonometric or exponential function with its corresponding Maclaurin series where
the number of terms in series follows discrete uniform distributions $\Uniform{3}{12}$ for UDTLZ2, UDTLZ4 and UDTLZ5, and $\Uniform{12}{19}$ in the case of UDTLZ3.

\section{Experimental Setup and Evaluations}\label{sec:experiments}
This section presents the results of evaluating the proposed reduced empirical distribution-based comparison operator as well as the techniques from~\cite{KMGT15}, \cite{KBT18}, and~\cite{MAG12} in the context of multi-objective optimization with uncertain objectives.
All comparison operators and the proposed UDTLZ benchmark have been integrated into the open-source Opt4J optimization framework~\cite{opt4jpaper} and can be paired with different optimization techniques.
The following experiments use the well-known \emph{Nondominated Sorting Genetic Algorithm II (NSGA-II)}~\cite{DAPM00} to optimize instances of UDTLZ as well as UZDT~\cite{KBT18} benchmarks.
NSGA-II is configured with a population size of $\lambda=25$ and performs optimization runs for 400 generations in each case.
The UZDT benchmark extends the well-known ZDT benchmark suite~\cite{ZDT00} and incorporates various uncertainties in the specification of all six bi-objective ZDT problems.
Also, the proposed UDTLZ problems are configured with $n=7$ decision variables and $m=3$ objective functions.

A sample size $N = 100$ is used for all instances of the UZDT and UDTLZ problems and for all comparison operators. 
This requires 100 iterative evaluations of each objective function of each candidate solution, and allows a good balance between the comparison accuracy and execution time. 
The proposed comparison operator uses a maximum of 20 steps in the approximated \acp{CDF}, while the histogram-based approach partitions each uncertainty distribution into intervals with a width of $\omega = 0.01$.
Both of these comparison operators are evaluated for a comparison threshold $\gamma = 0.7$ as suggested in~\cite{KBT18}.
Moreover, the three-stage approach is configured to use mean value in the average criterion and the 95\% quantile intervals in the spread criterion, as suggested in~\cite{KMGT15}.
Also, the threshold values $0.1$ and $0.3$ are used for these criteria, respectively.

\subsection{Convergence and Diversity}
\begin{figure*}
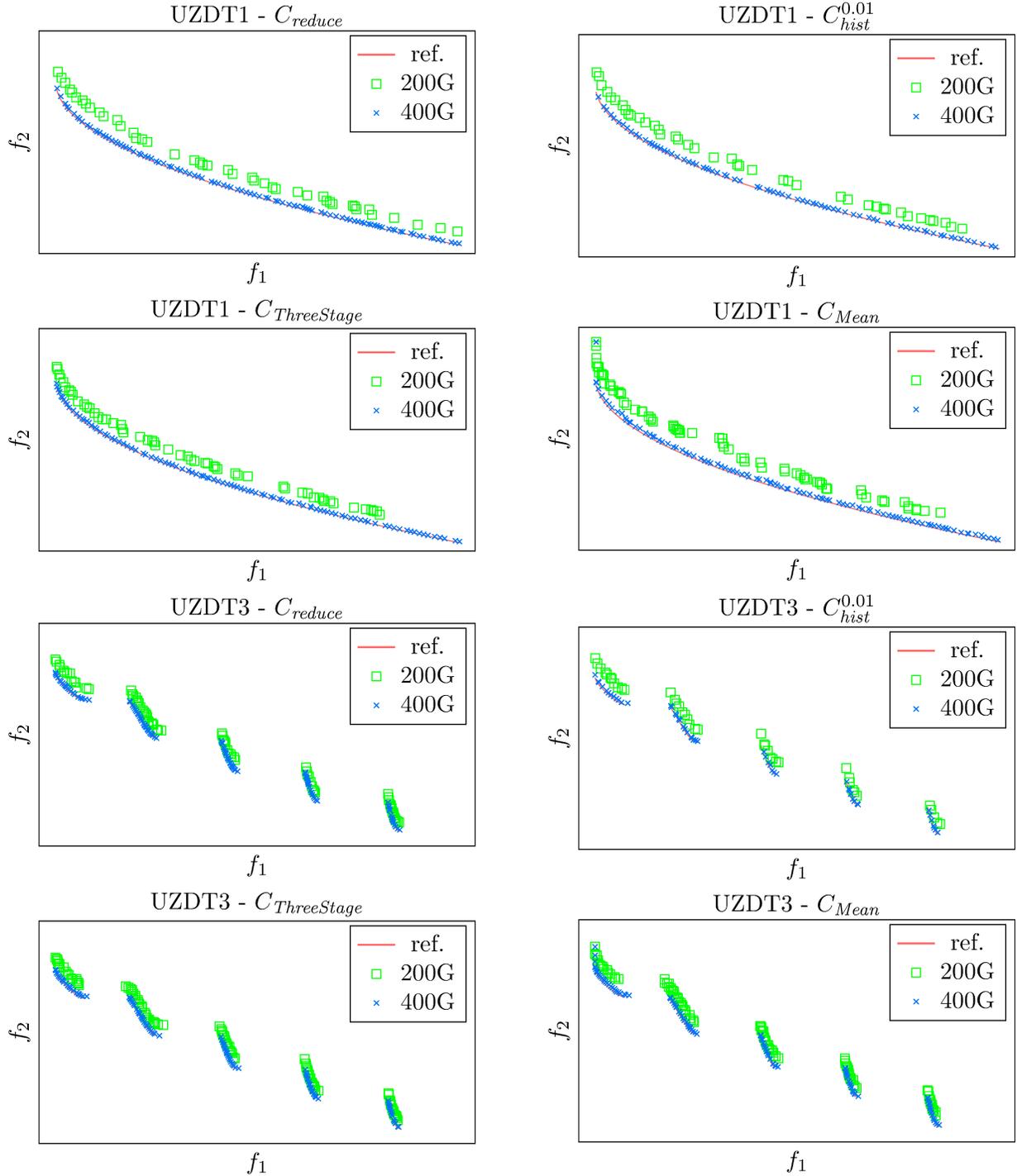

    \begin{subfigure}[]{0.52\textwidth}
    \frontsUZDTone{reduced}{8}{$\AReduced$}
    \end{subfigure}\hspace*{-.0\textwidth}
    \begin{subfigure}[]{0.52\textwidth}
    \frontsUZDTone{hist}{1}{$\AHistogram{0.01}$}
    \end{subfigure}\\[-1mm]
    \begin{subfigure}[]{0.52\textwidth}
    \frontsUZDTone{threestage}{7}{$\AThreeStage$}
    \end{subfigure}\hspace*{-.0\textwidth}
    \begin{subfigure}[]{0.52\textwidth}
    \frontsUZDTone{naive}{1}{$\AMeanBased$}
    \end{subfigure}\\[-1mm]
    \begin{subfigure}[]{0.52\textwidth}
    \frontsUZDTthree{reduced}{8}{$\AReduced$}
    \end{subfigure}\hspace*{-.0\textwidth}
    \begin{subfigure}[]{0.52\textwidth}
    \frontsUZDTthree{hist}{6}{$\AHistogram{0.01}$}
    \end{subfigure}\\[-1mm]
    \begin{subfigure}[]{0.52\textwidth}
    \frontsUZDTthree{threestage}{4}{$\AThreeStage$}
    \end{subfigure}\hspace*{-.0\textwidth}
    \begin{subfigure}[]{0.52\textwidth}
    \frontsUZDTthree{naive}{6}{$\AMeanBased$}
    \end{subfigure}
    \caption{Pareto front approximations for the UZDT1 and UZDT3 problems obtained after 200 and 400 generations of NSGA-II incorporating different comparison operators. Each case is run for 10 times and the median of the results is plotted. Shown are the mean values of the uncertain objectives $f_1$ and $f_2$.}
  \label{fig:pareto_one}
\end{figure*}

Figure~\ref{fig:pareto_one} depicts the Pareto front approximations for the UZDT1 and UZDT3 test problems~\cite{KBT18} obtained by NSGA-II incorporating the proposed $\AReduced$, the histogram-based $\AHistogram{0.01}$~\cite{KBT18}, the three-stage $\AThreeStage$~\cite{KMGT15} and the mean-based $\AMeanBased$~\cite{MAG12} comparison operators.
Shown are the reference Pareto front and two approximated fronts achieved after 200 and 400 generations considering only the mean value of the uncertain objectives $f_1$ and $f_2$. 
To obtain reliable results, each optimization run is repeated 10 times, the results are sorted with respect to the well-known $\epsilon$-dominance criterion~\cite{ZTLFD03}, and the run which provides the median $\epsilon$-dominance is chosen.  
The results show that, the first three comparison operators outperform the mean-based approach in most of the cases. 
However, the proposed comparison operator performs slightly better in converging to the Pareto front, while covering diverse regions in the objective space.

\begin{figure*}
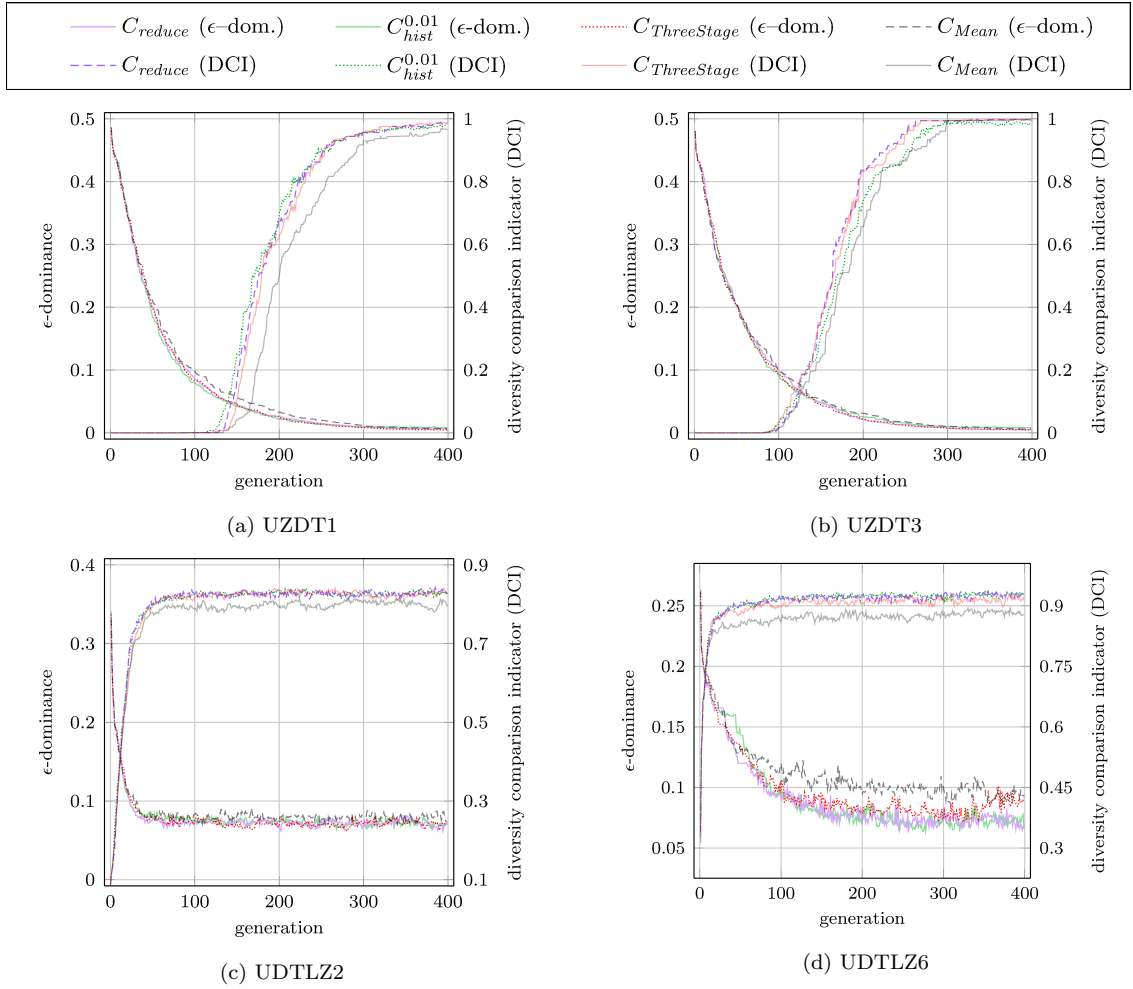

    \begin{center}
    \addlegends
    \vspace{0.3\baselineskip}
    
    \begin{subfigure}{0.42\textwidth}
    \dominanceDiversityUZDTone
    \caption{\label{fig:udtlz2}UZDT1}
    \end{subfigure}\qquad
    \begin{subfigure}{0.42\textwidth}
    \dominanceDiversityUZDTthree
    \caption{\label{fig:udtlz3}UZDT3}
    \end{subfigure}\\[0.2\baselineskip]
    \begin{subfigure}{0.42\textwidth}
    \dominanceDiversityUDTLZtwo
    \caption{\label{fig:udtlz5}UDTLZ2}
    \end{subfigure}\qquad
    \begin{subfigure}{0.42\textwidth}
    \dominanceDiversityUDTLZseven
    \caption{\label{fig:udtlz6}UDTLZ6}
    \end{subfigure}
    \end{center}
    \vspace*{-12pt}
    \caption{Resulting $\epsilon$-dominance and \ac{DCI} for the UZDT1, UZDT3, UDTLZ2 and UDTLZ6 problems. The results are obtained using optimizations that incorporate different comparison operators. For a reliable interpretation of the results, each case is repeated 10 times and at each generation, the median $\epsilon$-dominance and diversity values of all 10 runs are calculated.}
  \label{fig:dom_div}
  \vspace*{-10pt}
\end{figure*}

Figure~\ref{fig:dom_div} depicts the $\epsilon$-dominance~\cite{ZTLFD03} and \ac{DCI}~\cite{LYL14} calculated from the mean values of each uncertain objective for various instances of UZDT and UDTLZ problems.
Each result is calculated as the median of 10 optimization runs, calculated at each generation of NSGA-II.
$\epsilon$-dominance is calculated using the approach in~\cite{LGHT08} where the value of $\epsilon$ specifies how much the quality of candidate solutions found in an optimization run should be scaled so that these solutions weakly dominate the true Pareto front.
If the true Pareto front cannot be accurately described, which is the case for the UZDT and UDTLZ problems, it is replaced by a reference set of solutions. 
This reference set contains all solutions that turn out to be non-dominated after combining the approximated Pareto fronts of all runs of every optimization technique in the experiments.

To calculate \ac{DCI}, the objective space is partitioned into $m$-dimensional hypercubes of identical volume.
This allows to define a new coordinate system where solutions that lie within the same hypercube are treated as equal.
Given the coordinates of two sets of hypercubes that enclose the reference and an approximated Pareto front, 
the work in~\cite{LYL14} first calculates the sum of the Euclidean distances between each hypercube of the approximated front and the nearest hypercube of the reference front. 
Then it normalizes the result to a scale between 0 and 1 that respectively indicate the minimum and maximum possible degrees of diversity.   

Figure~\ref{fig:dom_div} shows the development of $\epsilon$-dominance and \ac{DCI} through the optimization, which indicates that the proposed comparison operator allows for a fast convergence to the Pareto front. 
The fluctuations in the results are due to the fact that the solutions are evolved with the help of elitism and crowding distance approaches~\cite{DAPM00} which do not differentiate candidate solutions in similar ways as in the post-optimization algorithms used to calculate $\epsilon$-dominance and \ac{DCI}.
Moreover, while the former approaches employ uncertainty-aware comparison operators which consider various characteristics for the uncertainty distribution of objective values, the latter algorithms always make decisions based on the mean values only.
It should be noted that the intensity of these fluctuations also depend on the extent of uncertainty in the objective functions such that for larger uncertainties, objective values of many candidate solutions would overlap.
This may affect the dominance relation between these candidate solutions and give rise to possible disagreements among the aforementioned approaches.

\subsection{Robustness}
\begin{figure*}[t]
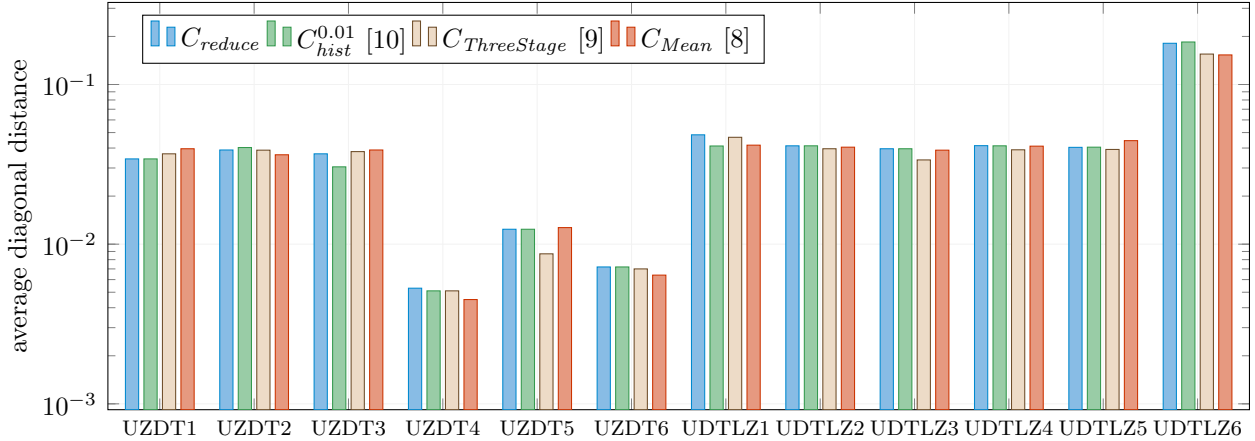

  \hspace*{-8pt}\distances
  \caption{Average diagonal distance of the solutions found for all UZDT and UDTLZ problems using optimizations that incorporate different comparison operators. Each case is averaged over the results of 10 independent optimization runs.}
    \vspace*{-3pt}
  \label{fig:robustness}
\end{figure*}

In the following, the proposed comparison operator $\AReduced$ is compared with the techniques in~\cite{KBT18},~\cite{KMGT15} and~\cite{MAG12} in terms of robustness of the found solutions for the UZDT and UDTLZ problems. 
The robustness is measured in this paper as the diagonal distance between worst- and best-case objective values such that a solution is said to be more robust if it delivers a smaller value for diagonal distance. 
Figure~\ref{fig:robustness} shows the results obtained for all evaluated comparison operators and all UZDT and UDTLZ problem instances.
In each case, the diagonal distance is calculated for all solutions in the Pareto front approximations of 10 optimization runs, and averaged. 

The results in Figure~\ref{fig:robustness} show that none of the comparison operators performs considerably better in finding robust solutions. 
This can be justified by the fact that the proposed, the histogram-based and the mean-based comparison operators do not consider the deviation of uncertain objective values as a comparison criterion.
The three-stage comparison operator takes this deviation into account in its third stage. 
However, we have observed that for all tested problems, in at least 94\% of cases, objective values can be distinguished in the first two stages, \ie the deviation criterion is used in less than 4\% of all comparisons. 
Also, in more than 92\% of the extended comparisons, the compared uncertain objective values are considered equal as their deviations are not noticeably different. 
This indicates that although a deviation criterion is employed, it is not likely to find solutions with significantly smaller uncertainties in their objective values.
Note that the diagonal distance does not reflect the accuracy of the comparison operators in distinguishing differently distributed uncertain objective values.

\section{Conclusion}\label{sec:conclusion}
Multi-objective optimization problems are often subject to various sources and forms of uncertainty in their decision variables and objective functions.
A wide range of approaches has been proposed in the literature to distinguish candidate solutions with objective values specified by confidence intervals, probability distributions or sampled data.
However, these approaches typically fail to treat uncertainties with non-standard distributions accurately and efficiently. 
In this paper, we investigated a variety of techniques that offer the comparison of uncertain objective values, and proposed novel techniques for the calculation of
the probability that an uncertain objective of one solution is more favorable than the same objective of the other solution.
To enable discovering robust solutions to problems with multiple uncertain objectives, we incorporated the proposed comparison operator into existing optimization techniques such as evolutionary algorithms. 
We also extended well-known multi-objective problems with various uncertainties and proposed a benchmark for evaluating robust optimization techniques. 
The proposed comparison operators and benchmark suite have been integrated into an existing optimization framework that provides a selection of multi-objective optimization problems and algorithms.
Experiments show that compared to existing techniques, our enhanced comparison operator achieves higher optimization quality and imposes lower overheads to the optimization process.

\bibliographystyle{alpha}
\bibliography{references}

\end{document}